\documentclass{article}
\DeclareUnicodeCharacter{FF0C}{,}

\usepackage{arxiv}
\usepackage[square,numbers]{natbib}
\usepackage[utf8]{inputenc}
\usepackage[T1]{fontenc}
\usepackage{hyperref}
\usepackage{url}
\usepackage{booktabs}
\usepackage{amsfonts}
\usepackage{nicefrac}
\usepackage{microtype}
\usepackage{lipsum}
\usepackage{graphicx}
\usepackage{threeparttable}
\usepackage{pdflscape}
\usepackage{doi}
\usepackage{verbatim}
\usepackage{mdframed}
\usepackage{fvextra}
\usepackage[english]{babel}
\usepackage{csquotes}
\usepackage{tabularx}
\usepackage{pifont}
\usepackage{subcaption}
\usepackage{amsmath}
\usepackage{longtable}
\newcolumntype{P}[1]{>{\raggedright\arraybackslash}p{#1}}
\usepackage{comment}
\usepackage[table]{xcolor}
\usepackage{multirow}
\usepackage{array}
\usepackage{svg}
\usepackage{diagbox}
\usepackage{amssymb}
\usepackage[normalem]{ulem}
\usepackage{algorithm}
\usepackage{algpseudocode}
\usepackage{listings}
\usepackage{fancyvrb}
\usepackage{minted}
\usepackage{makecell}

\algnewcommand\Input{\item[\textbf{Input:}]}
\algnewcommand\Output{\item[\textbf{Output:}]}

\usepackage{soul}
\usepackage{xcolor}

\title{Constructing Executable Analytical Knowledge Representations for Meta-Analysis Synthesis Using an Agentic Harness}

\author{
	\href{https://orcid.org/0009-0006-6918-4989}{\includegraphics[scale=0.06]{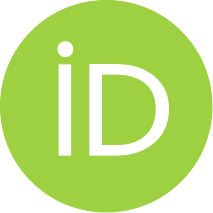}\hspace{1mm} Lingbo Li} \thanks{Corresponding author: L.Li5@massey.ac.nz} \\
	School of Mathematical and Computational Sciences\\
	Massey University\\
	Auckland, New Zealand \\
	\And
	\href{https://orcid.org/0000-0002-9124-2536}{\includegraphics[scale=0.06]{orcid.pdf}\hspace{1mm}Anuradha Mathrani} \\
	School of Mathematical and Computational Sciences\\
	Massey University\\
    Auckland, New Zealand\\
  	\And
    \href{https://orcid.org/0000-0001-9416-1435}{\includegraphics[scale=0.06]{orcid.pdf}\hspace{1mm}Teo ~Susnjak} \\
	School of Mathematical and Computational Sciences\\
	Massey University\\
	Auckland, New Zealand \\
}

\renewcommand{\shorttitle}{...}

\begin{document}
\maketitle

\begin{abstract}
Meta-analysis synthesis highlights a fundamental challenge in knowledge-based scientific analysis: structured evidence does not by itself represent the analytical knowledge required for executable computation. Decisions about evidence assignment, analytical contrasts, outcome and time-point alignment, effect-size formulation, and methodological admissibility must be made explicit before statistical execution. Existing automated approaches often embed these decisions in model outputs, generated code, or workflow traces rather than representing them as independently verifiable knowledge.
We introduce the Executable Analytical Knowledge Representation (EAKR), a machine-actionable representation of the knowledge required to transform structured study evidence into an executable meta-analysis. An EAKR represents participating evidence, analytical relations, numerical inputs, methodological constraints, provenance, and unresolved issues. We operationalise EAKR in MetaSynDec, an agentic harness in which large language models propose structured updates and deterministic services govern schema- and contract-based validation and execution.
Across 58 synthesis units, MetaSynDec constructed EAKRs for all units, and 57 proceeded to successful statistical execution. Of 56 units with sufficient published information to define a reference analysis object, 38 (67.9\%) achieved complete object fidelity and 42 (75.0\%) exact evidence-set agreement, with a mean Jaccard similarity of 0.909. Generated and published confidence intervals overlapped in 54 of 55 units (98.2\%). MetaSynDec outperformed direct LLM generation in reference synthesis-structure agreement (57/58 versus 23/58; p < 0.001) and, among 23 jointly completed units, exact reference-formulation agreement (23/23 versus 1/23; p < 0.001).
These findings provide controlled feasibility evidence that EAKR supports formal validation, traceability, statistical execution, and improved methodological agreement relative to direct LLM generation.
\end{abstract}

\keywords{Executable knowledge representation,
Agentic harness,
Large language models,
automated meta-analysis synthesis,
Scientific evidence synthesis}

\section{Introduction}
Structured data are increasingly underpinning artificial-intelligence systems for scientific analysis and decision support \cite{Ravi2022FAIRAIModels,Huerta2023FAIRAI}; however, machine-readable evidence alone is not necessarily analysis-ready in its original form \cite{Fernandes2023DataPreparation}. Before deterministic computation can be performed, structured evidence must be converted into task-specific analytical representation. This representation must capture both the available data and the analytical knowledge required to determine how individual records relate to the intended analysis and whether the requirements of the computational method have been satisfied \cite{Huerta2023FAIRAI,Fernandes2023DataPreparation}.
Without this intermediate knowledge layer, the reasoning from evidence to computation remains difficult to inspect, validate, reuse, and reproduce \cite{Huerta2023FAIRAI}.

\begin{figure}[htbp]
    \centering
    \includegraphics[width=\textwidth]{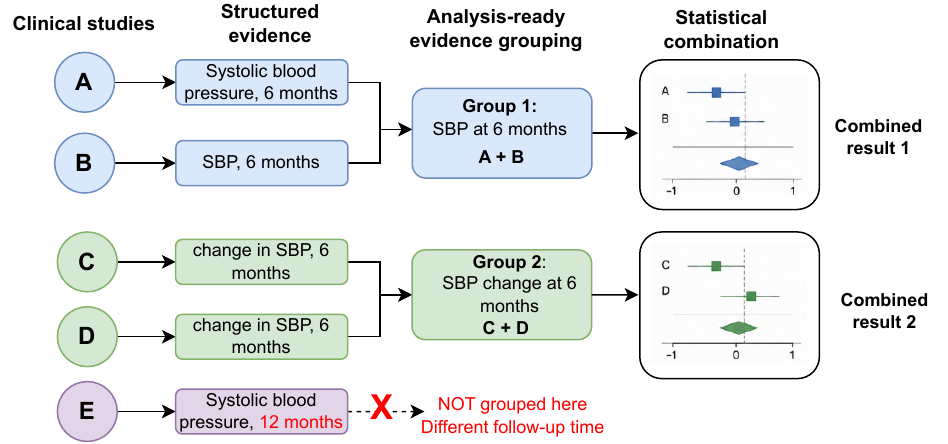}
    \caption{Meta-analysis synthesis as evidence routing before statistical combination. Clinical studies are first transformed into structured evidence records. Rather than being combined directly, the records are grouped according to the analytical question they answer (e.g., outcome and follow-up time). Only records within the same analysis-ready group are statistically combined, whereas incompatible records are routed to different analyses. This intermediate grouping is the focus of the proposed Executable Analytical Knowledge Representation (EAKR).}
    \label{fig:meta}
\end{figure}

The synthesis stage of meta-analysis illustrates this problem particularly clearly. It combines quantitative results from multiple studies addressing a common research question, for example, to estimate the overall effectiveness of an intervention across several trials \cite{cooper_research_2017,deeks2019meta}. Yet valid synthesis requires more than aggregating all available results. The selected results must represent sufficiently comparable populations, interventions, comparators, outcomes, measurement times, and effect estimates \cite{cooper_research_2017,egger_systematic_2008,hedges_statistical_2014}.
These requirements create an intermediate analytical task between evidence extraction and statistical pooling. For example, studies may report blood pressure outcomes referring to six-month systolic blood pressure (SBP), or changes in SBP from baseline, or twelve-month SBP. Although each result may have been extracted correctly, the records do not necessarily constitute a valid common synthesis (see Figure \ref{fig:meta}). An automated meta-analysis synthesis process must determine whether the outcomes represent the same construct, whether their measurement times are compatible, which arms serve as intervention and comparator, and whether the reported statistics support a common effect-size formulation. These determinations construct synthesis-specific knowledge about how the extracted records should be interpreted, related, and transformed for analysis.

Existing automation has primarily supported the stages on either side of this intermediate analytical task. Upstream, machine-learning and large language models (LLMs) methods can extract study characteristics and numerical results from publications \cite{marshall2019systematicreviewautomation,mutinda_automatic_2022,mutinda_autometa_2022,schmidt2025dataextraction}. However, the resulting structured records do not necessarily represent the synthesis-specific knowledge required to organise the extracted evidence for a particular meta-analysis. Downstream, statistical software and reproducible analysis pipelines can estimate predefined meta-analysis models, but typically assume that compatible contrasts and analysis-ready numerical inputs have already been specified \cite{choi_latent_2007,marot_moderated_2009,viechtbauer_conducting_2010,debray_framework_2019}. Some automated meta-analysis systems also support particular synthesis-stage operations, but these functions generally begin from evidence that has already been organised into an analysis-ready representation \cite{cumpston2021pico,mckenzie2019chapter9}. What remains comparatively less developed is support for constructing that representation from extracted study records. Such a representation must make explicit which evidence belongs to a common synthesis, how the evidence relates to the intended comparison and outcome, which analytical formulation is applicable, and whether the resulting inputs satisfy the requirements for statistical execution.

Constructing such a representation requires contextual interpretation as well as methodological validation. Recent advances in LLMs and agentic AI provide new mechanisms for supporting the interpretive part of this process \cite{Luo2024LLMReviews,Scherbakov2025LLMReviews}. LLMs may help determine whether differently named outcomes refer to the same construct, identify relevant measurement times, interpret intervention--comparator relationships, and relate study records to a synthesis objective. Existing LLM-based systems apply these interpretive capabilities directly within the synthesis process through task-specific assistance, code generation, and increasingly automated agentic workflows \cite{Reason2024AutomateNMA,manalyzer,metamind}. Language-model components may interpret extracted study records, determine which evidence should be combined, construct contrasts, select effect-size formulations, and generate or execute the corresponding analysis. These decisions are typically produced and propagated within the model-driven workflow rather than first being externalised as a dedicated analytical representation. As a result, the evidential basis, methodological constraints, and unresolved issues underlying the analysis may remain embedded in prompts, intermediate messages, generated code, tool calls, or final reports.

\begin{figure}[htbp]
    \centering
    \includegraphics[width=\textwidth]{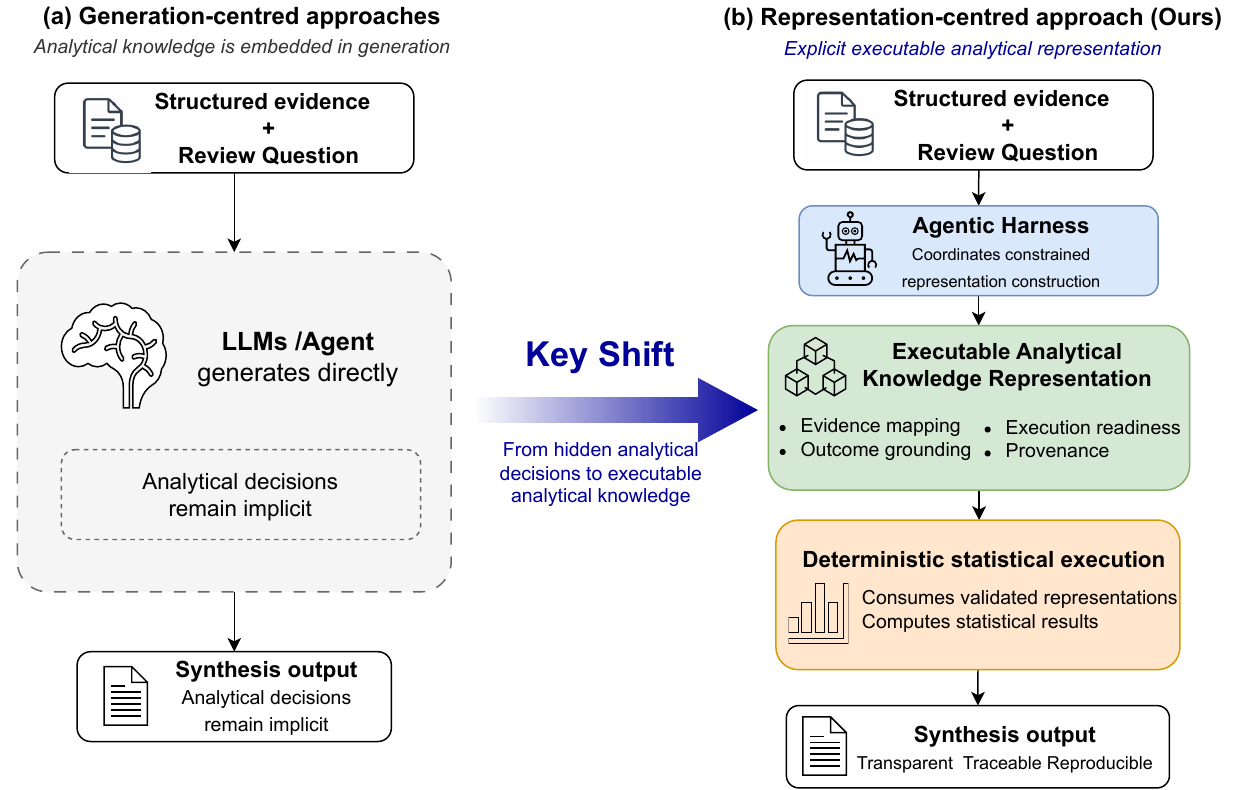}
    \caption{Conceptual shift from generation-centred reasoning to representation-centred reasoning. Existing approaches often embed analytical decisions inside prompts, intermediate reasoning, generated code, or final outputs. In contrast, the proposed approach constructs an explicit executable analytical representation before deterministic statistical execution. We implement this representation-centred workflow through an agentic harness that coordinates constrained reasoning and representation updates.}
    \label{fig:concept}
\end{figure}

We address this limitation by defining an Executable Analytical Knowledge Representation (EAKR), a representation of the synthesis knowledge required to transform structured study evidence into an executable meta-analysis (Figure \ref{fig:concept}). An EAKR identifies the participating evidence and specifies how that evidence relates to the synthesis objective by recording the analytical formulation, required numerical inputs, computational method, methodological constraints, provenance, and unresolved issues. It therefore represents not only the data to be analysed, but also the analytical claims and conditions under which their combination is considered valid. These claims and conditions are expressed in a form that allows the representation’s completeness, consistency, and methodological admissibility to be evaluated against formal constraints and method-specific input contracts before statistical computation is permitted. In this sense, an EAKR is executable not merely because it can be translated into statistical code, but because execution is contingent on the successful validation of the analytical knowledge from which that code is derived.
We specialise the EAKR for meta-analysis synthesis and instantiate it in a proposed agentic harness, that is, an orchestration framework that coordinates language-model components, validation mechanisms, constraint-checking mechanisms, and deterministic statistical services through bounded state transitions. Language-model components propose candidate interpretations and updates to the EAKR, while schemas, domain rules, methodological policies, completion criteria, and statistical input contracts determine whether those updates are admissible. Deterministic statistical services are invoked only when the resulting knowledge object satisfies the requirements of the intended analysis. The EAKR therefore serves as the central object through which the proposed agentic harness supports analytical reasoning, execution gating, provenance tracking, and discrepancy diagnosis.
The main contributions of this work are as follows:

\begin{itemize}
\item We formulate automated meta-analysis synthesis as a knowledge-construction problem between evidence extraction and statistical execution. This formulation distinguishes structured study records from the synthesis-specific analytical knowledge required to determine how those records should be interpreted, related, and transformed for analysis.
\item We define an Executable Analytical Knowledge Representation (EAKR), a  representation of the evidence, analytical relations, methodological choices, numerical requirements, constraints, provenance, and unresolved issues involved in a meta-analysis synthesis.
\item We develop an agentic harness for meta-analysis synthesis in which LLM-based interpretation and deterministic services collaboratively construct an EAKR through constrained state transitions. The resulting representation is checked for schema compliance, completeness, consistency, and methodological admissibility before statistical execution is permitted.
\item We empirically evaluate whether explicit knowledge construction improves synthesis execution, methodological agreement, numerical correctness, and run-to-run stability compared with direct LLM-based generation.
\end{itemize}

\section{Related work}
This section positions the proposed approach in relation to three bodies of work. We first examine conventional automation of meta-analysis synthesis, which largely assumes that an analysis has already been specified. We then consider LLM-based and agentic systems that can participate in constructing analyses but often leave their analytical decisions implicit. Finally, we draw on knowledge representation and intermediate-representation research to identify the requirements for making those decisions explicit, verifiable, and executable.

\subsection{Meta-analysis Synthesis Automation}
Meta-analysis software has traditionally automated statistical operations after reviewers have specified the analysis to be performed. Tools such as RevMan \cite{revman}, Comprehensive Meta-Analysis \cite{cma}, Stata \cite{Stata}, and SPSS \cite{spss} calculate study-level effects, fit fixed- or random-effects models, quantify heterogeneity, and generate numerical and graphical summaries. Their inputs, however, already encode decisions about eligible studies, intervention--comparator contrasts, outcomes, time points, effect measures, and required statistical variables.
Subsequent systems extended automation around this computational stage. Early approaches automated random-effects modelling and R-based analytical workflows \cite{viechtbauer_conducting_2010,michelson_automating_2014,yang_exploration_2018}. RevManHAL generated template-based synthesis text from structured RevMan outputs \cite{torres2017revmanhal}, while MetaCyto \cite{hu_metacyto_2018}, MetaMSD \cite{ryu_metamsd_2019}, and RICOPILI \cite{lam_ricopili_2020} supported scalable data integration and reproducible domain-specific analyses. Platforms such as CogTale \cite{sabates_cogtale_2021} and PsychOpen \cite{burgard_psychopen_2022} additionally supported standardised reporting, collaboration, and continuously updated synthesis outputs. These systems improve the execution, reproducibility, and dissemination of meta-analysis, but generally operate on a predefined analytical structure or within a workflow in which that structure is constrained in advance. 
Reviews \cite{khalil2022tools,scotti2025artificial,li2026transforming} of evidence-synthesis automation report a similar pattern: most tools address statistical computation or reporting, whereas fewer address the decisions that transform extracted evidence into a set of valid analyses.

The remaining problem is therefore not simply how to automate statistical pooling. It is how to computationally construct the analytical representation that pooling requires: which evidence belongs to each synthesis unit, which contrasts are valid, how outcomes and time points are aligned, and which numerical inputs satisfy the selected analytical formulation. Conventional synthesis automation largely begins after this representation has been established.

\subsection{LLM-Based and Agentic Evidence-Synthesis Systems}
LLMs extend evidence-synthesis automation by supporting tasks that require interpretation rather than only predefined computation. Systems such as LatteReview \cite{rouzrokh2025lattereview}, GREP-Agent \cite{hu2025enhancing}, and Otto-SR \cite{cao2025automation} support tasks including literature screening, relevance assessment, structured data extraction, and review orchestration. These systems primarily extract and structure study-level data rather than represent the analytical reasoning through which those data are transformed into a meta-analysis.

A smaller group of systems connects language models more directly to synthesis and statistical analysis. Empowering Meta-Analysis generates scientific summaries from included-study abstracts \cite{ahad2024empowering}. Reason et al.\ use GPT-4 to generate and execute code for network meta-analysis and to interpret comparative treatment effects \cite{Reason2024AutomateNMA}. Manalyzer proposes a multi-agent workflow spanning screening, extraction, statistical analysis, and report generation \cite{manalyzer}, while MetaMind combines semantic retrieval, multi-agent extraction, and generated statistical code for Bayesian network meta-analysis \cite{metamind}. These studies demonstrate that LLMs can connect extracted evidence to analytical procedures rather than merely supporting evidence collection.
Nevertheless, generating an analysis is not equivalent to explicitly representing its analytical basis. In existing workflows such as Manalyzer and MetaMind, synthesis-stage decisions are distributed across prompts, agent interactions, intermediate artefacts, and generated code. The resulting analysis may therefore be executable without providing a stable object that states why particular evidence was combined and whether every methodological requirement has been satisfied.

General-purpose agent frameworks such as AutoGen \cite{autogen} and LangGraph \cite{langgraph} do not by themselves address this representational requirement. Although they support agent coordination, tool use, state management, and workflow control, they do not define how study evidence should be interpreted and organised into a valid meta-analysis. They therefore provide the infrastructure for executing a synthesis workflow, but not the domain-specific representation needed to make its analytical decisions explicit and verifiable.

This distinction between workflow orchestration and analytical representation is particularly important in LLM-based synthesis. Although LLMs can participate in constructing an analysis, the decisions they generate through interpretation must be externalised before deterministic statistical execution. Otherwise, those decisions remain embedded in prompts, agent interactions, or generated code and cannot be independently inspected, verified against explicit constraints, revised, or traced to their supporting evidence. The unresolved gap is therefore not a lack of additional agents or code-generation capabilities, but the absence of a formally defined representation that connects evidence interpretation with statistical execution.

\subsection{Knowledge Representations for Analytical Reasoning}
Knowledge engineering provides a basis for defining such an interface. A central principle of knowledge representation is to encode domain entities, relationships, rules, and constraints separately from the procedures that operate on them \cite{davis1993knowledge}. This separation allows a system to inspect and validate what is known or asserted before using it to determine an action. 
Related forms of separation appear in other computational domains. Compiler intermediate representations preserve program semantics between source code and machine execution, enabling staged transformation, optimisation, and verification \cite{lattner2021mlir,fehr2025xdsl}. Scientific workflow representations describe computational steps, data dependencies, software environments, and provenance to support reproducible execution \cite{leo2024workflowrun,pritchard2025reproducibility}. Process-aware decision-support systems represent goals, conditions, states, and constraints so that proposed actions can be evaluated before they are performed \cite{leewis2024decision,seidel2023modelbased}. Although these representations serve different domains, they share a common architectural function: they make execution-relevant semantics explicit at the boundary between interpretation and action.

Existing representations in meta-analysis capture study evidence, planned methods, or analysis-ready data, but they do not explicitly represent how these elements are connected to form an executable analysis. Structured study records and outcome tables describe the evidence reported by individual studies, while analysis datasets contain numerical variables prepared for statistical software \cite{higgins2024cochrane}. Protocols and statistical scripts may additionally document planned methods and computational operations. Each representation captures part of the synthesis process, but none constitutes a unified analytical object that links evidence records to synthesis units, intervention--comparator contrasts, outcome and time-point interpretations, effect-size formulations, method-specific input requirements, and execution constraints.

The missing representation is therefore neither another evidence-extraction schema nor a generic workflow description. It must express the analytical commitments that connect structured evidence to a particular executable meta-analysis. It must identify the participating evidence, state how that evidence is interpreted and related, specify the intended analytical formulation, expose unresolved ambiguities, preserve provenance, and support validation against methodological and computational constraints. 
The present work addresses this representation gap through the Executable Analytical Knowledge Representation. The EAKR functions as an intermediate analytical object between structured study evidence and deterministic statistical execution. In doing so, it complements conventional synthesis software, which executes predefined analyses, and LLM-based systems, which can propose analytical decisions, by making those decisions explicit and verifiable before computation.

\section{Methods}

This section presents a representation-centred method for constructing executable analytical knowledge from structured study evidence. The method assumes as input a set of normalised study-level records describing study metadata, treatment and control arms, reported outcomes, measurement times, and numerical summaries, together with a review question. The objective is not to generate a meta-analysis result directly, but to construct an explicit computational representation of the analytical decisions required before statistical execution. We refer to this object as EAKR. MetaSynDec operationalises this method as an agentic harness. Language-model reasoning is used for interpretation-intensive decisions, while schemas, methodological constraints, deterministic checks, and statistical input contracts govern how the representation may be updated and when it may proceed to statistical execution. The following subsections first define the EAKR and its meta-analysis-specific state structure, then describe the agentic harness used to construct it, and finally present representative transition procedures for agent-assisted analytical planning and deterministic outcome typing.

\subsection{Executable Analytical Knowledge Representation for Meta-Analysis Synthesis} \label{subsec:EAKR}
The Executable Analytical Knowledge Representation (EAKR) is a structured computational representation of the analytical knowledge required to transform descriptive study evidence into a deterministic analytical disposition and, where feasible, an execution-ready specification for statistical synthesis. Extracted study records describe what individual studies reported, but they do not specify how those records should be interpreted, grouped, compared, or converted into valid statistical inputs. The EAKR makes these decisions explicit by representing analytical intent, evidence grounding, execution requirements, provenance, validation status, unresolved ambiguity, and whether the available evidence supports quantitative pooling. During synthesis, the EAKR is constructed incrementally as evidence is interpreted and organised, with proposed updates checked against predefined schemas, methodological constraints, and statistical input contracts before being incorporated into the analysis specification.

Let \(X=\{x_1,\ldots,x_n\}\) denote a set of structured study records and let \(Q\) denote the review question. A direct mapping \((X,Q)\rightarrow Y\) from structured evidence to statistical output \(Y\) bypasses the intermediate analytical knowledge required to establish that the computation is valid. We therefore formulate synthesis as the incremental construction of an EAKR:
\begin{equation}
(X,Q)\rightarrow S_0 \rightarrow S_1 \rightarrow \cdots \rightarrow S_K=S^\ast \rightarrow Y,
\label{eq:representation_construction}
\end{equation}
where \(S_0\) is the initial EAKR derived from the review question and structured evidence, \(S_i\) is the representation after construction stage \(i\), \(S^\ast\) is the schema- and contract-valid analysis-ready EAKR, and \(Y\) is the resulting analytical output. Depending on the available evidence, \(Y\) may contain either a pooled statistical result or a methodologically admissible non-pooling decision, together with its supporting rationale. Each transition \(S_{i+1}=T_i(S_i)\) modifies only a predefined part of the representation and must satisfy the constraints associated with that construction stage.

In the present meta-analysis instantiation, the EAKR is maintained as the synthesis state:
\begin{equation}
S_i=\langle C,E,O_i,U_i,A_i\rangle,
\label{eq:synthesis_state}
\end{equation}
where \(C\) represents the review context, \(E\) the structured study evidence, \(O_i\) the outcome knowledge constructed up to stage \(i\), \(U_i\) the analysis objects constructed up to stage \(i\), and \(A_i\) the accumulated provenance, validation, and unresolved-issue records. The synthesis state provides the concrete computational structure through which the EAKR is updated during workflow execution. Each transformation \(T_i\) operates on \(S_i\), updating only its designated components while preserving the remaining state. The final state \(S_K=S^\ast\) represents an analysis-ready EAKR that satisfies the applicable schema and input-contract constraints. Details of the EAKR components are provided in Table~\ref{tab:eakr_components}.

The components of the synthesis state represent persistent analytical knowledge rather than one-to-one workflow stages. In particular, \(O_i\) is constructed incrementally: outcome classification and type are established during outcome interpretation, whereas measurement scale, unit harmonisation, and numerical-input readiness are added or updated as outcome-specific evidence is matched and organised during synthesis planning. The resulting outcome knowledge is subsequently used to construct the analysis objects in \(U_i\). Thus, \(O_i\) records the interpreted and standardised properties of the available outcome evidence, whereas \(U_i\) specifies how that evidence is assembled into executable analysis objects.

{
\begin{table}[htbp]
\centering
\caption{Meta-analysis-specific components of the Executable Analytical Knowledge Representation.}
\label{tab:eakr_components}
\begin{tabular}{p{1.2cm}p{3.1cm}p{7.8cm}}
\toprule
\textbf{Component} & \textbf{Role} & \textbf{Information represented} \\
\midrule
\(C\) & Review context &
Population, intervention, comparator, target outcomes, intended measurement times, and review-level analytical constraints. \\

\(E\) & Structured evidence &
Study identifiers, study arms, reported outcomes, measurement times, numerical summaries, and source-record identifiers. \\

\(O\) & Outcome knowledge &
Outcome classification, outcome type, measurement-scale and unit
harmonisation, and numerical-input readiness. \\

\(U\) & Analysis objects &
Evidence-to-analysis mappings, included study sets, treatment--control assignments, measurement-time groupings, analytical disposition, selected statistical form, and required inputs or non-pooling rationale. \\

\(A\) & Provenance and control information &
Source links, validation results, warnings, rejected updates, unresolved ambiguities, and execution-readiness status. \\
\bottomrule
\end{tabular}
\end{table}
}

An analysis object \(u \in U\) represents one planned analysis and is defined as
\begin{equation}
u
=
\left\langle
I_u,
P_u,
Y_u,
T_u,
F_u,
D_u,
M_u,
R_u,
Q_u
\right\rangle,
\label{eq:analysis_object}
\end{equation}
where \(I_u\) is a non-empty set of evidence records selected for the analysis; \(P_u\) specifies the intervention--comparator contrast; \(Y_u\) identifies the target outcome inherited from the corresponding outcome representation in \(O\); \(T_u\) specifies the time-point selection strategy; \(F_u\) specifies the analytical formulation, such as endpoint values or change scores; \(D_u\) specifies the numerical data form and whether the selected evidence provides, or can be transformed into, the inputs required for that form; \(M_u\) specifies the analytical disposition, effect measure, and statistical model; \(R_u\) contains provenance records supporting the analytical specification; and \(Q_u\) contains unresolved ambiguity, conflict, or missing information that may prevent execution.

Each analysis object has an analytical disposition
\(\delta_u \in \{\mathtt{POOL},\mathtt{NOT\_POOLED}\}\)
encoded within \(M_u\). The \(\mathtt{POOL}\) disposition indicates that the selected evidence satisfies the predefined requirements for quantitative synthesis, whereas \(\mathtt{NOT\_POOLED}\) indicates that pooling is not supported. A \(\mathtt{NOT\_POOLED}\) disposition constitutes a completed analytical decision rather than a construction failure.
The disposition is assigned after evidence has been selected according to the specified outcome, contrast, and time-point strategy. Excluded evidence and the corresponding reasons are recorded in \(R_u\). Missing, conflicting, or ambiguous information is recorded in \(Q_u\); if such an issue could affect the validity of the analysis, it is classified as blocking, and no final disposition may be assigned until it is resolved. Otherwise, the analysis object is assigned either \(\mathtt{POOL}\) or \(\mathtt{NOT\_POOLED}\).

For an evidence record \(e\), eligibility for inclusion in \(u\) is defined as
\begin{equation}
\begin{aligned}
\mathrm{Eligible}(e\mid u)
={}&
\mathrm{OutcomeCompat}(e\mid Y_u)
\land
\mathrm{ContrastCompat}(e\mid P_u)\\
&\land
\mathrm{TimeSelected}(e\mid T_u)
\land
\mathrm{DataCompat}(e\mid F_u,D_u,M_u).
\end{aligned}
\label{eq:evidence_eligibility}
\end{equation}

\(\mathrm{OutcomeCompat}(e\mid Y_u)\) requires the normalised outcome concept of \(e\) to match \(Y_u\), with compatible scale, direction, and statistical form, unless an explicitly registered transformation is available. \(\mathrm{ContrastCompat}(e\mid P_u)\) requires the intervention and comparator roles in \(e\) to agree with the contrast specified by \(P_u\). Any reversal of treatment roles or outcome direction must be represented explicitly and recorded in \(R_u\). \(\mathrm{DataCompat}(e\mid F_u,D_u,M_u)\) requires the reported numerical fields to support the analytical formulation specified by \(F_u\) and the effect-size method specified by \(M_u\), either directly or through an admissible transformation. For an analysis object assigned \(\delta_u = \mathtt{NOT\_POOLED}\), this predicate determines whether the evidence is analytically characterisable but does not require sufficient inputs for pooled estimation.

The time-selection predicate depends on the strategy encoded by \(T_u\):
\begin{equation}
\mathrm{TimeSelected}(e\mid T_u)=
\begin{cases}
\mathbb{I}\!\left[t_e=\max \mathcal{T}_{s(e)}^{\mathrm{eligible}}\right], & T_u=\mathtt{FINAL},\\
\mathbb{I}\!\left[G_{T_u}(t_e)=\tau_u\right], & T_u=\mathtt{GROUPED},
\end{cases}
\label{eq:time_selection}
\end{equation}
where \(t_e\) is the follow-up time associated with evidence record \(e\), \(\mathcal{T}_{s(e)}^{\mathrm{eligible}}\) is the set of eligible follow-up times reported by the corresponding study, \(G_{T_u}\) is the grouping function constructed by the planning stage from the available follow-up measurements, and \(\tau_u\) is the time group represented by analysis object \(u\). Thus, the final-follow-up strategy selects one eligible measurement from each study without requiring identical measurement times, whereas the grouped strategy assigns measurements to analysis objects according to their automatically generated time groups.

An analysis object may proceed to deterministic analytical execution only when it satisfies
\begin{equation}
\begin{aligned}
\mathrm{Ready}(u)
={}&
\mathrm{ValidOutcome}(u)
\land
\mathrm{ValidMapping}(u)
\land
\mathrm{ValidContrast}(u)\\
&\land
\mathrm{ValidInput}(u)
\land
\mathrm{ValidProvenance}(u)
\land
\mathrm{Resolved}(u).
\end{aligned}
\label{eq:ready_predicate}
\end{equation}
\(\mathrm{ValidOutcome}(u)\) requires a specified outcome concept, measurement scale, direction, and compatible statistical form. \(\mathrm{ValidMapping}(u)\) requires every evidence record included in \(I_u\) to satisfy Eq.~\eqref{eq:evidence_eligibility}. \(\mathrm{ValidContrast}(u)\) requires consistent treatment and control roles across the included studies. \(\mathrm{ValidInput}(u)\) requires the numerical fields needed by the selected statistical method to be available or computable through an admissible transformation. \(\mathrm{ValidProvenance}(u)\) requires each inclusion, exclusion, transformation, time-point selection, and material planning decision to be linked to its supporting evidence, rule, or reviewer action. Finally,
\(\operatorname{Resolved}(u)\iff Q_u^{\operatorname{blocking}}=\varnothing\)
so an analysis object containing unresolved blocking ambiguity cannot proceed to execution.

The readiness predicate is an analytical execution contract. It establishes that an analysis object is internally consistent, sufficiently specified, provenance-complete, and assigned a valid analytical disposition. 
Once an analysis object satisfies Eq.~\eqref{eq:ready_predicate}, deterministic execution determines the appropriate analytical outcome according to the encoded statistical specification. Where the predefined requirements for quantitative synthesis are satisfied, the workflow proceeds to pooled statistical analysis. Otherwise, the analysis object records a non-pooled disposition that satisfies the applicable methodological and input-contract constraints, together with the eligible evidence, exclusion rationale, and supporting provenance, without attempting to compute a pooled estimate. If \(\mathrm{Ready}(u)=0\), the missing, conflicting, or ambiguous elements are recorded in \(Q_u\) and the associated provenance log \(A\), and the analysis object is withheld from execution.
This formulation separates three computational layers: descriptive study evidence, executable analytical knowledge, and deterministic analytical execution. The next subsection describes how MetaSynDec constructs and updates the EAKR through an agentic harness.

\subsection{Constructing Executable Analytical Knowledge Representations with an Agentic Harness}
MetaSynDec is an agentic harness that implements the incremental construction and refinement of EAKRs. Figure~\ref{fig:architecture} illustrates its architecture. The framework comprises four principal components: a global orchestrator, stage controllers, a shared synthesis state, and stage-local services. In the current implementation, the shared synthesis state is defined by Eq.~\ref{eq:synthesis_state}. The global orchestrator determines the sequence of construction stages, while each stage controller governs a specific transformation of the current synthesis state.
Formally, each stage controller implements a constrained transition \(T_i\). It may access and modify only the state fields authorised for that stage, invokes the relevant stage-local services, and validates the proposed update against the applicable schema and methodological constraints. A transition from \(S_i\) to \(S_{i+1}\) is committed only when these conditions are satisfied; otherwise, the current state is preserved and the validation failure or unresolved issue is recorded in the synthesis state. Consequently, intermediate synthesis decisions are represented as explicit and inspectable changes to the EAKR that are checked against applicable schemas and methodological constraints, rather than being propagated solely through free-form messages, prompts, or generated text.

{
\begin{figure}[htbp]
    \centering
    \includegraphics[width=\linewidth]{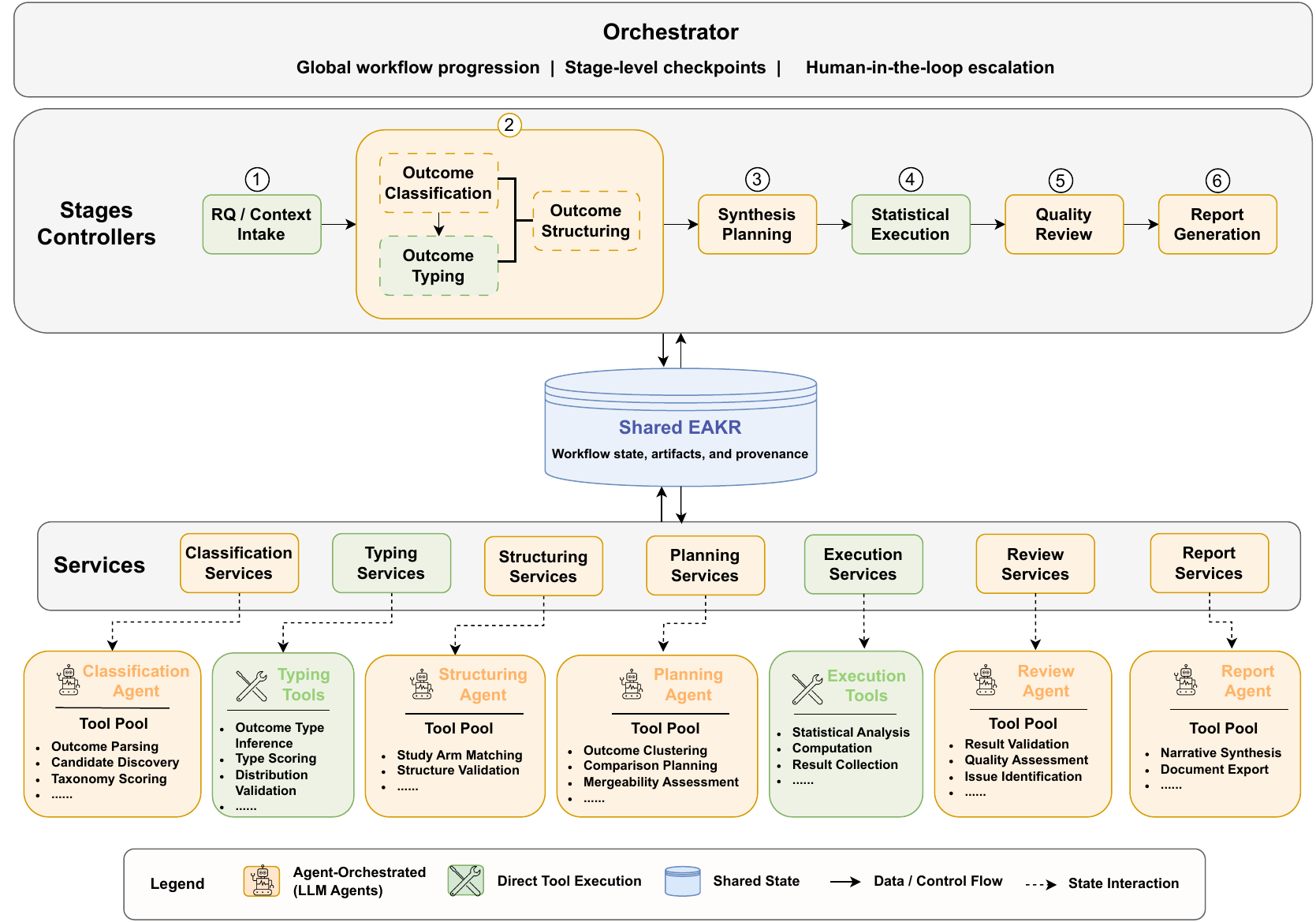}
    \caption{MetaSynDec agentic harness for constructing Executable Analytical Knowledge Representations. The harness coordinates a global orchestrator, stage controllers, a shared EAKR, and stage-local services. The review-context stage initialises the analytical question and target comparison. Outcome-processing stages interpret and structure reported study outcomes, while synthesis planning groups compatible evidence, assigns treatment--control roles, and constructs executable analysis objects. Deterministic statistical services operate only on analysis objects that satisfy the required readiness conditions. Quality review records inconsistencies and unresolved issues, and report generation produces outputs grounded in the final representation and its provenance. Solid arrows indicate control and data flow; dashed arrows indicate interaction with the shared representation.}
    \label{fig:architecture}
\end{figure}
}

\paragraph{\textbf{Architecture components.}}
The \textit{\textbf{Orchestrator}} controls the overall construction sequence by selecting the next stage, managing checkpoints, and handling retry, termination, and escalation outcomes. It does not perform synthesis reasoning directly. Each \textit{\textbf{Stage Controller}} implements one stage-specific EAKR transition: it invokes the relevant services, verifies whether the proposed update satisfies the stage contract, and returns a routing outcome to the orchestrator. The \textit{\textbf{Shared EAKR}} serves as the persistent analytical workspace and stores the review context, structured evidence, interpreted outcome knowledge, executable analysis objects, statistical input requirements, provenance records, validation results, and unresolved issues. Stage-local \textit{\textbf{Services}} perform the operations required by each transition, including deterministic procedures for rule-based typing, validation, and statistical execution, and LLM-assisted procedures for interpretation-intensive tasks.
Table \ref{tab:state_transition} links the architecture to the representation-construction formulation introduced in Section \ref{subsec:EAKR}. Each stage modifies only the predefined EAKR fields required for downstream processing. The construction sequence therefore makes explicit how outcome interpretation, statistical typing, evidence grouping, contrast assignment, measurement-time handling, and execution readiness are progressively added to the representation.

{
\begin{table}[htbp]
\centering
\caption{Stage-wise construction of the Executable Analytical Knowledge Representation in MetaSynDec.}
\label{tab:state_transition}
\begin{tabular}{p{0.17\linewidth} p{0.23\linewidth} p{0.29\linewidth} p{0.22\linewidth}}
\hline
\textbf{Stage controller} &
\textbf{Input EAKR fields} &
\textbf{Knowledge operation} &
\textbf{EAKR update} \\
\hline

RQ / context intake &
\((X,Q)\) &
Record the review question, target comparison, and review-level constraints &
\(C\): population, treatment, control, target outcomes, eligibility context \\

Outcome classification &
\((C,E)\) &
Map reported outcomes to clinically meaningful categories and candidate groupings &
Outcome-category labels, candidate groupings, taxonomy links \\

Outcome typing &
\((C,E)\) and classified outcomes &
Assign the statistical outcome type&
\(O\): outcome typed \\

Outcome structuring &
\((C,E,O)\) &
Convert study-level outcome records into normalised, analysis-ready evidence objects &
Structured outcome records, arm links, time information, validation issues \\

Synthesis planning &
\((C,E,O)\) &
Group compatible evidence and construct executable analysis objects &
\(U\): evidence mappings, study sets, treatment--control roles, time groups, planned statistical forms \\

Statistical execution &
\((C,E,O,U)\) with \(\mathrm{Ready}(u)=1\) &
Construct statistical inputs and execute deterministic analysis routines &
Statistical results, confidence intervals, heterogeneity measures, execution logs, forest plots \\

Quality review &
Full EAKR and statistical results &
Check consistency, completeness, provenance, and unresolved issues &
\(A\): validation status, warnings, issue records, escalation flags \\

Report generation &
Execution-ready EAKR and statistical results &
Generate a representation-grounded synthesis narrative&
\(Y\): narrative synthesis, provenance-linked outputs \\
\hline
\end{tabular}
\end{table}
}

\paragraph{\textbf{Inspectability of analytical decisions.}}
MetaSynDec makes intermediate analytical decisions inspectable by storing them as explicit records within the EAKR. Each record identifies the decision type, affected evidence records, selected value, supporting provenance, producing stage, validation status, and unresolved-issue flags. For example, an outcome-mapping record identifies which study-level measurements are assigned to an analysis object, together with their reported labels, measurement times, scales, study identifiers, and comparability checks. A treatment-role record identifies which arms are assigned as treatment and control for each study and links those assignments to the original structured evidence.
This representation allows a downstream statistical result to be traced to the analytical decisions and evidence records that produced it. The present study evaluates this property through qualitative disagreement localisation, using localised disagreements to demonstrate how model behaviour can be inspected and traced.

\paragraph{\textbf{Stage-bounded agentic reasoning.}}
MetaSynDec uses bounded agentic reasoning to construct the EAKR incrementally. The global construction sequence remains under deterministic control of the orchestrator, while LLM-based reasoning is invoked only within stages that require contextual interpretation. A stage-local agent proposes an update to the permitted EAKR fields, but it cannot directly modify unrelated fields or bypass the validation contract.

For each agent-assisted stage \(i\), the bounded agent is defined as
\begin{equation}
B_i=
\left\langle
F_i^{\mathrm{read}},
F_i^{\mathrm{write}},
\mathcal{T}_i,
K_i,
V_i,
C_i
\right\rangle,
\label{eq:stage_bounded_agent}
\end{equation}
where \(F_i^{\mathrm{read}}\) and \(F_i^{\mathrm{write}}\) define the readable and writable representation fields, \(\mathcal{T}_i\) defines the stage-local tool set, \(K_i\) is the maximum number of within-stage reasoning--tool iterations, \(V_i\) defines the validation rules for proposed updates, and \(C_i\) defines the stage completion condition. In the implementation used for evaluation, deterministic stages are executed once \((K_i=1)\), whereas each agent-assisted stage is allowed at most three stage-local update attempts \((K_i=3)\). Each attempt may include one reasoning step, one or more admissible tool calls, and validation of the proposed EAKR update. Agent reasoning is therefore bounded by the active stage, permitted representation fields, available tools, iteration limit, and validation requirements. Table~\ref{tab:tool_pool} summarises the stage-local services, their representation operation, and their execution mode.

{
\begin{table}[htbp]
\centering
\caption{Stage-local services used to construct and execute the EAKR.}
\label{tab:tool_pool}
\begin{tabular}{p{0.18\linewidth} p{0.31\linewidth} p{0.32\linewidth} p{0.14\linewidth}}
\hline
\textbf{Stage / service} &
\textbf{Main tools} &
\textbf{Representation operation} &
\textbf{Execution mode} \\
\hline

RQ / context intake &
Context parser; schema checker &
Initialise review context and analytical constraints &
Deterministic \\

Outcome classification &
Outcome parser; candidate discovery; taxonomy scorer &
Propose clinically meaningful outcome categories and groupings &
Stage-local agent \\

Outcome typing &
Type rules; compatibility scorer; distribution validator &
Assign statistical outcome type &
Deterministic \\

Outcome structuring &
Arm matcher; time normaliser; structure validator &
Construct normalised evidence objects and identify structural issues &
Stage-local agent \\

Synthesis planning &
Outcome grouper; evidence mapper; contrast planner; comparability assessor &
Construct executable analysis objects and map evidence to them &
Stage-local agent \\

Statistical execution &
Input-contract checker; effect calculator; statistical runner; result collator &
Consume execution-ready analysis objects and compute statistical results &
Deterministic \\

Quality review &
Consistency checker; result validator; issue detector &
Record representation inconsistencies, warnings, and unresolved issues &
Stage-local agent \\

Report generation &
Representation-grounded generator; provenance linker; document exporter &
Generate outputs grounded in the final EAKR and statistical results &
Stage-local agent \\
\hline
\end{tabular}
\end{table}
}

\paragraph{\textbf{Escalation and control handling.}}

If a stage does not satisfy its completion condition within the allowed iteration bound, the orchestrator does not permit the workflow to proceed with an incomplete representation. It records the failed checks, unresolved fields, rejected updates, and tool traces in the audit component \(A\), and marks the affected condition as unresolved in the corresponding \(Q_u\).
When the unresolved condition can potentially be addressed through another admissible stage-local action, the stage may be retried within the predefined iteration bound. If the bound is exhausted and the issue requires reviewer judgement, the workflow enters \texttt{WAITING\_FOR\_HUMAN} and presents the unresolved condition together with its supporting evidence and validation trace.
The reviewer may select one of three resolution actions: \textit{retry}, \textit{continue with the current state}, or \textit{exclude}. A \textit{retry} decision authorises an additional automated attempt without directly changing the analytical content of the EAKR. A \textit{continue} decision records that the current condition is accepted as non-blocking and stores the reviewer decision and rationale in \(A\) and \(R_u\). An \textit{exclude} decision removes the affected evidence record or candidate analysis object from further construction and records the exclusion and its rationale in the corresponding evidence mapping and provenance records.
After any reviewer action, the affected analysis object is revalidated against Eq.~\eqref{eq:ready_predicate}. The workflow proceeds only when all blocking conditions have been resolved. If no admissible automated or reviewer-authorised resolution produces an execution-ready representation, the workflow terminates with status \texttt{FAILED}; the unresolved condition remains recorded in \(Q_u\), and the affected analysis object is withheld from deterministic statistical execution.

\subsection{Representative Procedures for Constructing and Updating EAKRs} \label{sec:disagreement_analysis}
MetaSynDec uses two classes of procedures to construct and update EAKRs. Deterministic procedures are used when an update can be derived from structured fields, explicit rules, or statistical input contracts. Agent-assisted procedures are used when the update requires contextual interpretation or synthesis judgement. This subsection presents one representative example of each class: synthesis planning as a stage-bounded agent-assisted EAKR update, and outcome typing as a deterministic EAKR update.

\paragraph{\textbf{Agent-assisted synthesis planning.}}
Synthesis planning transforms interpreted outcome knowledge into executable analysis objects within the EAKR. In terms of the meta-analysis-specific representation defined in Eq.\eqref{eq:synthesis_state}, the procedure updates \(U\) with evidence-to-analysis mappings, study inclusion sets, intervention--comparator assignments, measurement-time groupings, planned statistical forms, and analysis-ready input rows. It also updates \(A\) with provenance, validation results, unresolved issues, and execution traces. An analysis object is passed to deterministic statistical execution only when it satisfies the readiness predicate in Eq.~\eqref{eq:ready_predicate}.

Operationally, the planning stage is entered after outcome structuring and is executed only when the workflow state is \texttt{STRUCTURED}. The stage controller constructs a planning request from the current EAKR and delegates the construction process to a goal-directed agent. Within each planning attempt, the agent evaluates the current representation, identifies unresolved analytical requirements, and selects admissible operations from the stage-local tool pool. These operations include agent-assisted semantic decisions, such as intervention--comparator role assignment and contrast construction. 
Algorithm~\ref{alg:planning} summarises this bounded agent-based construction process. Each iteration of the algorithm represents one complete planning attempt rather than a single tool call. The resulting candidate analysis objects are schema- and contract-checked before being committed to \(U\), while associated issues, artifacts, and action traces are recorded in \(A\). The stage returns \texttt{PLANNED} when all analysis objects satisfy the readiness predicate. In the evaluated implementation, the agent was permitted at most \(K_{\mathrm{plan}}=3\) planning attempts. Human escalation was considered only after all three attempts had been exhausted; otherwise, the stage terminated with \texttt{FAILED}.

{
\begin{algorithm}[t]
\caption{Agent-based EAKR construction for synthesis planning}
\label{alg:planning}
\begin{algorithmic}[1]

\Require Current EAKR
\(S_i=\langle C,E,O,U,A\rangle\),
planning policy \(P\),
stage-local tool pool \(\mathcal{T}_{\mathrm{plan}}\),
maximum planning attempts \(K_{\mathrm{plan}}\)

\Ensure Updated EAKR \(S_{i+1}\) and routing status \(\rho \in \{\mathtt{PLANNED}, \mathtt{WAITING\_FOR\_HUMAN}, \mathtt{FAILED}\}\)

\State Construct planning request \(R_{\mathrm{plan}}\) from
\(C,E,O,U,A\)

\For{planning attempt \(k=1\) to \(K_{\mathrm{plan}}\)}

    \State Agent evaluates the current EAKR and identifies unresolved
    planning requirements

    \State Agent selects an admissible sequence of operations
    \(\Pi_k\subseteq\mathcal{T}_{\mathrm{plan}}\)

    \State Execute \(\Pi_k\) to construct candidate analysis objects
    \(\widehat{U}_k\)

    \State Validate \(\widehat{U}_k\) against the planning-stage
    contract

    \State Append issues, artifacts, action traces, and validation
    results to \(A\)

    \If{\(\widehat{U}_k\) is valid}
        \State \(U\leftarrow\widehat{U}_k\)
    \EndIf

    \If{all analysis objects satisfy
    Eq.~\eqref{eq:ready_predicate}}
        \State Set workflow state to \texttt{PLANNED}
        \State \(S_{i+1}\leftarrow\langle C,E,O,U,A\rangle\)
        \State \Return \(S_{i+1},\mathtt{PLANNED}\)
    \EndIf

    \State Update \(R_{\mathrm{plan}}\) from the current EAKR

\EndFor

\If{the remaining unresolved issues require human judgement}
    \State Record a structured human-input request in \(A\)
    \State Set workflow state to \texttt{WAITING\_FOR\_HUMAN}
    \State \(\rho\leftarrow\mathtt{WAITING\_FOR\_HUMAN}\)
\Else
    \State Record a terminal planning failure in \(A\)
    \State Set workflow state to \texttt{FAILED}
    \State \(\rho\leftarrow\mathtt{FAILED}\)
\EndIf

\State \(S_{i+1}\leftarrow\langle C,E,O,U,A\rangle\)
\State \Return \(S_{i+1},\rho\)

\end{algorithmic}
\end{algorithm}
}

\paragraph{\textbf{Deterministic outcome typing.}}
In contrast to agent-assisted synthesis planning, outcome typing is implemented as a deterministic EAKR update to the outcome-knowledge component \(O\) in Eq.~\eqref{eq:synthesis_state}. The typing service processes the set of structured outcome records produced by the evidence-classification stage and assigns each record one of four statistical data-family labels:
\textit{continuous},
\textit{binary},
\textit{time-to-event}, or
\textit{ambiguous}.
Continuous outcomes are typically represented by numerical summaries such as means and standard deviations or standard errors. For example, a study arm may report \texttt{mean = 4.2}, \texttt{SD = 1.3}, and \texttt{n = 50}. Binary outcomes are typically represented by event counts together with the corresponding group sizes, such as \texttt{events = 12} and \texttt{n = 50}. Time-to-event outcomes are typically represented by survival-analysis statistics, such as a hazard ratio together with its confidence interval or standard error, for example, \texttt{HR = 0.72} and \texttt{95\% CI = [0.55, 0.94]}. When the available structured evidence does not provide sufficient support for one unique data family, the record is labelled \textit{ambiguous}.
Outcome typing relies exclusively on structured statistical evidence. The runtime evaluates a fixed set of discriminative evidence signals using globally defined weights to infer the statistical data family of each outcome record. The resulting label serves as a data-type constraint for downstream outcome structuring.

Before assigning a data-family label, the runtime identifies the structured outcome records
\(
E_{\mathrm{out}}\subseteq E
\)
from the classified evidence stored in the EAKR evidence component. Each outcome record represents a single structured statistical observation (e.g., a mean, standard deviation, event count, or hazard ratio) associated with a study-level outcome, and preserves its associated study identifier and source path. The typing procedure is first applied independently to each record. Records associated with the same outcome within the same study are subsequently grouped to determine a unified study-level outcome type.

For each outcome record \(e\in E_{\mathrm{out}}\), the typing service extracts a discriminative evidence-signal vector
\begin{equation}
\mathbf{z}_{\mathrm{disc}}(e)=\big(z_{\mathrm{tte}}(e),z_{\mathrm{binary}}(e),z_{\mathrm{central}}(e),z_{\mathrm{dispersion}}(e),z_{\mathrm{uncertainty}}(e),z_{\mathrm{unit}}(e)\big),
\label{eq:evidence_vector}
\end{equation}
Each component is binary and indicates whether the corresponding category of structured statistical evidence is present in the record. 
The time-to-event signal \(z_{\mathrm{tte}}(e)\) is activated by structured fields associated with hazard ratios, survival probabilities, survival times, Cox proportional-hazards models, or log-rank analyses. The binary signal \(z_{\mathrm{binary}}(e)\) is activated by fields representing event counts, non-event counts, cases, deaths, responses, odds ratios, risk ratios, or related dichotomous quantities. These signals are treated as family-specific gates because they provide direct evidence of the underlying statistical data family.
The central-value signal \(z_{\mathrm{central}}(e)\) is activated by means, averages, changes, differences, deltas, or related continuous summaries; the dispersion signal \(z_{\mathrm{dispersion}}(e)\) by standard deviations, variances, interquartile ranges, ranges, or related variability measures; the uncertainty signal \(z_{\mathrm{uncertainty}}(e)\) by standard errors, confidence-interval bounds, confidence levels, or related measures of estimation uncertainty; and the unit signal \(z_{\mathrm{unit}}(e)\) by the presence of an explicit measurement unit. 
Distinguishing dispersion from uncertainty prevents generic uncertainty fields from being treated as direct evidence of a continuous family. Dispersion provides stronger continuous-family support because they describe how observed values vary across measured outcome, whereas uncertainty is not specific to any one outcome family, since they may be reported for continuous, binary, or time-to-event estimates. Unit fields may support a continuous or binary interpretation, but a unit alone is not sufficient to assign the outcome family.

The runtime also records a contextual bounded-value indicator
\(z_{\mathrm{bounded}}(e)
=
\mathbb{I}\!\left[
\exists x\in V(e): 0\le x\le 1
\right]\),
where \(V(e)\) denotes the set of numeric values appearing in \(e\). Values in \([0,1]\) frequently observed in binary outcomes expressed as proportions or probabilities, but may also occur in continuous measures such as utility scores, normalized indices, or other bounded scales.
The indicator is therefore retained as contextual evidence in the inference trace rather than treated as a discriminative typing signal. It does not contribute directly to the family support scores, but remains available to support the interpretation of ambiguous record-level classifications and their subsequent aggregation at the study-outcome level.

Outcome typing follows a hierarchical deterministic inference strategy. First, family-specific signals are evaluated. If \(z_{\mathrm{tte}}(e)=1\), the evidence item is assigned to the \textit{time-to-event} family. Otherwise, if \(z_{\mathrm{binary}}(e)=1\), it is assigned to the \textit{binary} family. These signals are treated as deterministic gates because they correspond to statistical representations that are more family-specific than generic summary-value fields. Weighted evidence aggregation is used only when neither gate is triggered.
For each candidate label \(r \in \{\textit{continuous}, \textit{binary}, \textit{time-to-event}\}\), the support score is computed as
\begin{equation}
s_r(e)=\sum_j \omega_{r,j}z_j(e),
\label{eq:typing_score}
\end{equation}
where \(j\) indexes the evidence signals in Eq.~\eqref{eq:evidence_vector}, and \(\omega_{r,j}\) is a fixed evidence-strength weight. The weights encode predefined evidence strength. The complete weight matrix is reported in Table \ref{tab:typing_weights}.
When weighted aggregation is required, the record-level label is assigned as
\begin{equation}
r^\ast(e)
=
\begin{cases}
\displaystyle \arg\max_r s_r(e),
& \text{if the maximum support score is unique and positive},\\[5pt]
\textit{ambiguous},
& \text{otherwise}.
\end{cases}
\label{eq:assigned_family}
\end{equation}
An outcome record is therefore labelled \textit{ambiguous} when multiple candidate families share the maximum support score or when all candidate families receive zero support. 

After record-level typing, records associated with the same study-level outcome are grouped to determine a unified outcome type. Let
\(
G_o
\)
denote the set of records associated with study-level outcome \(o\), and let
\(
R_o
=
\left\{
r^\ast(e)
:
e\in G_o,\;
r^\ast(e)\neq\textit{ambiguous}
\right\}
\)
denote the set of non-ambiguous record-level labels observed within that group. The aggregated outcome type is defined as
\begin{equation}
r^\ast(o)
=
\begin{cases}
r,
& \text{if } R_o=\{r\},\\[3pt]
\textit{ambiguous},
& \text{otherwise}.
\end{cases}
\label{eq:aggregated_outcome_type}
\end{equation}
Thus, ambiguous records do not override consistent family-specific evidence from other records belonging to the same study outcome. However, if no non-ambiguous record is available, or if the records provide conflicting family labels, the study-level outcome remains \textit{ambiguous}. The record-level scores, assigned labels, bounded-value indicators, and aggregation decision are retained in the inference trace stored in \(A\). Algorithm~\ref{alg:typing} summarises the complete deterministic outcome-typing update.

{
\begin{table}[htbp]
\centering
\caption{Fixed evidence-strength weights used for deterministic outcome typing.}
\label{tab:typing_weights}
\begin{tabular}{lccc}
\toprule
Evidence signal
& Continuous
& Binary
& Time-to-event \\
\midrule
Time-to-event signal \(z_{\mathrm{tte}}\)
& 0 & 0 & 3 \\

Binary signal \(z_{\mathrm{binary}}\)
& 0 & 3 & 0 \\

Central-value signal \(z_{\mathrm{central}}\)
& 2 & 0 & 0 \\

Dispersion signal \(z_{\mathrm{dispersion}}\)
& 2 & 0 & 0 \\

Uncertainty signal \(z_{\mathrm{uncertainty}}\)
& 1 & 1 & 1 \\

Measurement-unit signal \(z_{\mathrm{unit}}\)
& 1 & 1 & 0 \\

\bottomrule
\end{tabular}
\begin{tablenotes}
\footnotesize
\item \textit{Note:} Weights denote predefined evidence strength: decisive (3), strong (2), and weak (1). 
\end{tablenotes}
\end{table}
}

{
\begin{algorithm}[htbp]
\caption{Deterministic EAKR construction of outcome typing and aggregation}
\label{alg:typing}
\begin{algorithmic}[1]

\Require EAKR \(S_i=\langle C,E,O,U,A\rangle\), fixed weight matrix \(\Omega\)
\Ensure Updated EAKR \(S_{i+1}=\langle C,E,O',U,A'\rangle\)

\State Define \(\mathcal{R}\leftarrow\{\textit{continuous},\textit{binary},\textit{time-to-event}\}\)
\State \(E_{\mathrm{out}}\leftarrow\) structured outcome records in \(E\)
\State \(\mathcal{G}\leftarrow\operatorname{GroupByStudyOutcome}(E_{\mathrm{out}})\)
\State \(O'\leftarrow O,\;A'\leftarrow A\)
\State Record \(\mathcal{G}\) and its grouping provenance in \(A'\)

\ForAll{\((o,G_o)\in\mathcal{G}\)}
    \ForAll{\(e\in G_o\)}

        \State Extract \(\mathbf{z}_{\mathrm{disc}}(e)\leftarrow
        (z_{\mathrm{tte}}(e),z_{\mathrm{binary}}(e),z_{\mathrm{central}}(e),
        z_{\mathrm{dispersion}}(e),z_{\mathrm{uncertainty}}(e),z_{\mathrm{unit}}(e))\)

        \State Extract the contextual bounded-value indicator \(z_{\mathrm{bounded}}(e)\)

        \ForAll{\(r\in\mathcal{R}\)}
            \State \(s_r(e)\leftarrow
            \sum_{j\in\mathcal{J}_{\mathrm{disc}}}\omega_{r,j}z_j(e)\)
        \EndFor

        \If{\(z_{\mathrm{tte}}(e)=1\)}
            \State \(r^\ast(e)\leftarrow\textit{time-to-event}\)
            \State \(\operatorname{rule}(e)\leftarrow\texttt{decisive\_tte}\)

        \ElsIf{\(z_{\mathrm{binary}}(e)=1\)}
            \State \(r^\ast(e)\leftarrow\textit{binary}\)
            \State \(\operatorname{rule}(e)\leftarrow\texttt{decisive\_binary}\)

        \Else
            \State \(s_{\max}(e)\leftarrow\max_{r\in\mathcal{R}}s_r(e)\)
            \State \(M(e)\leftarrow
            \{r\in\mathcal{R}:s_r(e)=s_{\max}(e)\}\)

            \If{\(s_{\max}(e)>0\) \textbf{and} \(|M(e)|=1\)}
                \State \(r^\ast(e)\leftarrow\) the unique element of \(M(e)\)
                \State \(\operatorname{rule}(e)\leftarrow
                \texttt{unique\_weighted\_maximum}\)
            \Else
                \State \(r^\ast(e)\leftarrow\textit{ambiguous}\)
                \State \(\operatorname{rule}(e)\leftarrow
                \texttt{ambiguous\_weighted\_result}\)
            \EndIf
        \EndIf

        \State \(A'\leftarrow A'\cup
        \{(e,\mathbf{z}_{\mathrm{disc}}(e),z_{\mathrm{bounded}}(e),
        \{s_r(e)\}_{r\in\mathcal{R}},r^\ast(e),\operatorname{rule}(e))\}\)

    \EndFor

    \State \(R_o\leftarrow
    \{r^\ast(e):e\in G_o,\;r^\ast(e)\neq\textit{ambiguous}\}\)

    \If{\(|R_o|=1\)}
        \State \(r^\ast(o)\leftarrow\) the unique element of \(R_o\)
        \State \(\operatorname{aggregation}(o)\leftarrow
        \texttt{consistent\_record\_evidence}\)
    \Else
        \State \(r^\ast(o)\leftarrow\textit{ambiguous}\)

        \If{\(|R_o|=0\)}
            \State \(\operatorname{aggregation}(o)\leftarrow
            \texttt{no\_resolved\_record}\)
        \Else
            \State \(\operatorname{aggregation}(o)\leftarrow
            \texttt{conflicting\_record\_families}\)
        \EndIf
    \EndIf

    \State \(O'[o]\leftarrow r^\ast(o)\)
    \State \(A'\leftarrow A'\cup
    \{(o,\{r^\ast(e):e\in G_o\},R_o,r^\ast(o),
    \operatorname{aggregation}(o))\}\)

\EndFor

\State \Return \(S_{i+1}=\langle C,E,O',U,A'\rangle\)

\end{algorithmic}
\end{algorithm}
}

\section{Experiments}
The experiments examine whether MetaSynDec can construct execution-ready EAKRs from structured evidence and whether those representations can support reliable meta-analysis execution. The evaluation is organised around the EAKR components defined in Eq.~\eqref{eq:synthesis_state} and the analysis-object structure specified in Eq.~\eqref{eq:analysis_object}.
The experiments consider construction effectiveness, representation and publication fidelity, provenance and intervention requirements, and robustness across review contexts and repeated executions. The comparison with direct generation is treated as a system-level ablation: it compares the integrated EAKR-centred MetaSynDec system with direct synthesis-plan generation.

\subsection{Datasets}
We evaluated MetaSynDec on the complete structured meta-analysis dataset reported in our previously published extraction study \cite{li_dataextraction}. All six available meta-analysis cases were included, comprising 58 randomised controlled trials, 58 predefined synthesis tasks, and 932 structured evidence records. A synthesis task corresponds to one execution-ready analysis object defined in the EAKR. Randomised trials are represented as normalised JSON study objects containing study metadata, participant characteristics, intervention and comparator descriptions, outcome measurements, measurement times, study-design attributes, and eligibility information.
Collectively, these study objects instantiate the structured-evidence component \(E\). From this evidence, MetaSynDec constructs the outcome-knowledge component \(O\), executable analysis objects \(U\), and the provenance and control component \(A\). The six cases were descriptively grouped by synthesis complexity based on the number of synthesis targets and the extent of outcome branching, measurement-time branching, multi-arm ambiguity, terminology variation, and required data transformation. Table \ref{tab:dataset_complexity} summarises the cases and their complexity characteristics.

{
%\begin{landscape}
\begin{table}[htbp]
\centering
\caption{Evaluation dataset grouped by synthesis complexity.}
\label{tab:dataset_complexity}
\setlength{\tabcolsep}{5pt}
% \begin{tabularx}{\textwidth}{l l l p{2cm} c l l p{4cm}}
\begin{tabularx}{\linewidth}{
    p{0.5cm}
    l
    c
    p{1.5cm}
    c
    p{1.5cm}
    p{1.5cm}
    X
}
\toprule
\textbf{MA} &
\textbf{Field} &
\textbf{RCTs} &
\textbf{Review-level outcomes} &
\textbf{Timepoints} &
\textbf{Synthesis units} &
\textbf{Extracted outcome records} &
\textbf{Complexity source} \\
\midrule

\multicolumn{8}{l}{\textit{Low synthesis complexity}} \\[1pt]
MA1 & Hypertension & 13 & 1 & 1 & 1 & 26 & Single outcome and single timepoint \\
MA4 & Diabetes mellitus & 6 & 1 & 1 & 1 & 22 & Limited multi-arm contrast ambiguity \\[4pt]

\multicolumn{8}{l}{\textit{Moderate synthesis complexity}} \\[1pt]
MA2 & Hypertension & 10 & 9 & 1 & 9 & 160 & Nested outcome hierarchy and multiple synthesis targets \\
MA3 & Diabetes mellitus & 10 & 5 & 1 & 5 & 140 & Nested outcome hierarchy and multiple synthesis targets \\
MA5 & Orthopedic & 7 & 7 & 1 & 7 & 147 & Outcome terminology variation \\[4pt]

\multicolumn{8}{l}{\textit{High synthesis complexity}} \\[1pt]
MA6 & Orthopedic & 12 & 14 & 2--3 & 35 & 437 & Nested outcome hierarchy with outcome terminology variation and timepoint branching \\
\midrule
\textbf{Total} & -- & \textbf{58} & \textbf{37} & -- & \textbf{58} & \textbf{932} & -- \\
\bottomrule
\end{tabularx}

\vspace{2mm}
\begin{minipage}{0.98\linewidth}
\footnotesize
\textit{Note.} Review-level outcomes denote distinct outcome concepts identified in the source meta-analyses, whereas synthesis units denote independent analytical targets after contrast, and measurement-time branching. Extracted outcome records refer to the structured study-level outcome entries supplied to the synthesis workflow. Complexity groups were assigned descriptively according to the number and interaction of outcome, contrast, measurement-time, and terminology-alignment decisions required for synthesis construction.
\end{minipage}
\end{table}
%\end{landscape}
}

\subsection{Evaluation Metrics} \label{sec:metrics}

The evaluation metrics were organised according to the four experimental questions (EQs). Results are reported for the primary execution, using the synthesis unit as the evaluation unit. The four EQs jointly evaluate the constructed EAKR at the representation, execution, and system levels. EQ1 evaluates whether the complete synthesis state reaches a schema- and contract-valid analytical outcome. EQ2 evaluates the semantic and analytical content represented in \(O\) and \(U\), together with the statistical outputs produced from that content. EQ3 evaluates the provenance and control information represented in \(A\), \(R_u\), and \(Q_u\), as well as robustness across meta-analysis contexts and repeated executions. EQ4 evaluates the integrated value of the complete EAKR-centred system relative to direct generation. The context and structured-evidence components \(C\) and \(E\) are treated as controlled experimental inputs; their preservation and source linkage are checked as part of input integrity and provenance evaluation.

\paragraph{\textbf{EQ1: Construction effectiveness.}}
\textit{Can MetaSynDec construct schema- and contract-valid EAKRs for the predefined synthesis units and carry eligible analytical objects through the complete synthesis workflow?}

Construction effectiveness evaluates MetaSynDec's ability to produce execution-ready analysis representations at the synthesis-unit level. A synthesis unit is considered execution-ready only when all required analysis objects satisfy Eq.~\eqref{eq:ready_predicate}
Workflow completion is evaluated separately. A synthesis unit is classified as workflow-complete only when quantitative statistical execution and all subsequent workflow stages have produced their required outputs. Under the current implementation, a unit with a final supported non-pooling disposition is counted as successful EAKR construction but not as full workflow completion, because it does not produce a statistical result for the downstream result-processing stages.
For each synthesis unit, we also record whether it became ready for execution within the allowed number of automated attempts or following a reviewer-authorised action. 
We therefore report the numbers and proportions of synthesis units that: (i) produce a schema- and contract-valid EAKR; (ii) produce a quantitative pooled statistical result; (iii) reach a final supported non-pooling disposition; (iv) complete the full workflow; (v) require reviewer-authorised action; and (vi) terminate without a valid analytical outcome.

\paragraph{\textbf{EQ2: Analytical-knowledge and result fidelity.}}
\textit{Do the constructed EAKRs faithfully represent the outcome knowledge, analytical decisions, and statistical results of the reference syntheses?}

Following the construction evaluation in EQ1, EQ2 assesses the fidelity of the resulting EAKRs. The evaluation examines whether the outcome knowledge represented in \(O\), the analytical decisions encoded in \(U\), and the statistical outputs generated from those decisions agree with the corresponding reference syntheses. 

For an evaluation object \(x\) and categorical field \(f\), exact agreement between the constructed and reference values is defined as
\begin{equation}
A_{x,f}
=
\mathbb{I}
\left[
\widehat{f}_x
=
f_x^{\mathrm{ref}}
\right],
\label{eq:field_agreement}
\end{equation}
where \(\widehat{f}_x\) denotes the value of field \(f\) constructed by MetaSynDec for evaluation object \(x\), and \(f_x^{\mathrm{ref}}\) denotes the corresponding reference value. Evaluation objects for which the reference field cannot be recovered unambiguously are excluded from the denominator for that field.

\textbf{Outcome-knowledge fidelity} evaluates whether the semantic and standardisation decisions stored in \(O\) agree with the corresponding reference synthesis. Outcome classification, outcome type, and measurement-scale and unit harmonisation are evaluated over the distinct review-level outcomes using the exact field-agreement indicator in Eq.~\eqref{eq:field_agreement}.
Outcome-classification agreement requires the constructed review-level outcome category to match the reference category. Outcome-type agreement requires the constructed type to match the reference label among \textit{continuous}, \textit{binary}, \textit{time-to-event}, and \textit{ambiguous}. Measurement-scale and unit-harmonisation agreement requires the identified scales and units, compatibility decisions, harmonised target unit, and any required conversions to match the reference representation.
Complete outcome-knowledge fidelity is defined as
\begin{equation}
A_o^{O}
=
A_{o,c}A_{o,t}A_{o,h},
\label{eq:complete_outcome_fidelity}
\end{equation}
where \(c\), \(t\), and \(h\) denote outcome classification, outcome type, and measurement-scale and unit harmonisation, respectively. Thus, \(A_o^{O}=1\) only when all three evaluated outcome-knowledge attributes agree with the reference representation.

Outcome classification, outcome type, measurement-scale and unit harmonisation, and semantic outcome fidelity are reported as the numbers and proportions of eligible distinct outcomes with exact agreement. Numerical-input readiness is represented in \(O\) but is not treated as a reference-agreement metric because the published meta-analyses did not consistently report pre-execution readiness states or transformation pathways. Its operational role is evaluated through the readiness contract under EQ1, whereas the provenance of any required transformations is evaluated under EQ3.

\textbf{Analysis-object fidelity} evaluates whether the analytical decisions encoded in \(U\) agree with the corresponding reference synthesis. The evaluated fields are evidence selection \(I_u\), intervention--comparator contrast \(P_u\), measurement-time selection \(T_u\), analytical formulation \(F_u\), effect measure \(M_u^{e}\), and statistical
model policy \(M_u^{m}\). The target-outcome field \(Y_u\) is not evaluated separately because it links the analysis object directly to the outcome knowledge in \(O\) and does not introduce an additional classification decision. The numerical data from \(D_u\) was also not evaluated as a standalone field. Its sufficiency formed part of EAKR validation and was reflected in successful analytical execution in EQ1, while its downstream adequacy was assessed through result fidelity. Provenance \(R_u\) and unresolved-issue records \(Q_u\) are evaluated under EQ3. 

Contrast-assignment agreement requires the intervention and comparator arms, including their orientation, to match the reference contrast. 
Measurement-time agreement requires the selected time point or time group, together with the applicable time-selection strategy, to match the reference synthesis. 
Analytical formulation agreement requires the analysis object to use the same analytical input form as the reference analysis, such as endpoint values or change scores. 
Effect-measure agreement requires the constructed analysis object to use the same effect measure as the reference synthesis (e.g. mean difference or standardised mean difference for a continuous outcome). 
Statistical model-policy agreement requires the same meta-analytic model policy, such as fixed-effect or random-effects. These categorical fields are evaluated using Eq.~\eqref{eq:field_agreement}.

Because exact evidence-set agreement does not distinguish between small and large set differences, partial evidence overlap is additionally quantified using the Jaccard coefficient:
\begin{equation}
J_{u,I}
=
\frac{
\left|\widehat{I}_u\cap I_u^{\mathrm{ref}}\right|
}{
\left|\widehat{I}_u\cup I_u^{\mathrm{ref}}\right|
}.
\label{eq:inclusion_jaccard}
\end{equation}
Exact evidence-set agreement occurs when \(\widehat{I}_u = I_u^{\mathrm{ref}}\), which is equivalent to \(J_{u,I}=1\). 

Complete analysis-object fidelity is evaluated only for analysis objects for which unambiguous reference values are available for all predefined analysis-object fields.
Let
\(\mathcal{F}^{U}=\{I,P,T,F,M^{e},M^{m}\}\)
For an eligible analysis object \(u\),
\begin{equation}
A_u^{U}
=
\prod_{f \in \mathcal{F}^{U}}
A_{u,f}.
\label{eq:complete_analysis_object_fidelity}
\end{equation}
Thus, \(A_u^{U}=1\) only when all six evaluated analytical decisions agree with the reference synthesis. Analysis objects lacking an unambiguous reference for any field in \(\mathcal{F}^{U}\) are excluded from the denominator of this composite metric.

\textbf{Result fidelity} evaluates whether the statistical outputs generated from the constructed analysis objects agree with the corresponding published results. It is assessed only for synthesis units that produce a quantitative result and have a directly comparable reference. 

Categorical result fidelity is evaluated at three complementary levels: direction-of-effect agreement, significance-decision agreement, and joint interpretation agreement. Unless otherwise specified, categorical agreement is evaluated according to Eq.~\eqref{eq:field_agreement}.
Direction-of-effect agreement assesses whether the generated and published estimates favoured the same group, without considering statistical significance. Significance-decision agreement assesses whether both estimates reached, or did not reach, statistical significance, without considering effect direction. Joint interpretation agreement requires the generated and published results to have the same combined direction-and-significance classification (e.g. statistically significant benefit, non-significant benefit, statistically significant harm, or non-significant harm). Direction and significance are therefore treated as component decisions, whereas joint interpretation is the stricter composite criterion.

Among them, significance-decision agreement is defined as
\begin{equation}
A_i^{\mathrm{sig}}
=
\mathbb{I}
\left[
\mathbb{I}\!\left(p_i^{M} < 0.05\right)
=
\mathbb{I}\!\left(p_i^{P} < 0.05\right)
\right].
\label{eq:significance_agreement}
\end{equation}

When pooled estimates were available in a directly comparable form, numerical fidelity was assessed using two complementary metrics: confidence interval overlap and confidence-interval-scaled deviation. These metrics were chosen to enable comparisons across synthesis units involving outcomes reported on heterogeneous scales and in different physical units.
Raw absolute differences between effect estimates were not aggregated across synthesis units. Although an absolute difference could be calculated for each unit, it would remain dependent on the outcome-specific scale and unit of measurement. Averaging or taking the median of deviations expressed in units such as mmHg, biochemical concentrations, clinical scores, bone density, and degrees of joint motion would combine non-commensurable quantities and would therefore not produce an interpretable cross-outcome measure of error. Relative effect-estimate error was likewise not used as an aggregate metric because it is undefined when the published estimate is zero and may become unstable when the published estimate is close to zero. Numerical fidelity was therefore evaluated using confidence-interval-based metrics, which allow meaningful comparison across heterogeneous synthesis units.

Confidence-interval overlap is defined as
\begin{equation}
A_i^{\mathrm{CI}}
=
\mathbb{I}
\left[
\max\left(l_i^{M}, l_i^{P}\right)
\leq
\min\left(u_i^{M}, u_i^{P}\right)
\right],
\label{eq:ci_overlap}
\end{equation}
where \(l_i^{M}\) and \(u_i^{M}\) denote the MetaSynDec confidence interval, and \(l_i^{P}\) and \(u_i^{P}\) denote the published confidence interval.

Confidence-interval-scaled deviation is defined as
\begin{equation}
D_i
=
\frac{
\left|
\theta_i^{M} - \theta_i^{P}
\right|
}{
\left(u_i^{P} - l_i^{P}\right)/2
}.
\label{eq:estimate_deviation}
\end{equation}
where \(\theta_i^{M}\) and \(\theta_i^{P}\) denote the MetaSynDec and published pooled-effect estimates, respectively. This dimensionless quantity expresses the point-estimate discrepancy relative to the half-width of the published confidence interval. A value of \(D_i=1\) indicates that the point-estimate difference equals one half-width of the published confidence interval, whereas smaller values indicate closer agreement relative to the uncertainty of the reference estimate.

\paragraph{\textbf{EQ3: Traceability and robustness.}} 
\textit{Does the constructed EAKR preserve sufficient provenance and control information to support traceable and robust execution?}

EQ3 evaluates whether the analytical decisions represented in the EAKR remain traceable to their supporting evidence and construction context, and whether these decisions remain stable across repeated executions. Traceability is evaluated through the provenance and control information represented in \(A\), \(R_u\), and \(Q_u\), whereas robustness is evaluated through repeated executions of each synthesis unit.

\textbf{Decision-level traceability} evaluates whether every analytical decision made during EAKR construction is explicitly recorded and linked to its supporting evidence, decision context, and subsequent workflow state. The evaluated records include source-record identifiers, evidence-inclusion and exclusion decisions, intervention--comparator assignments, outcome and time-point selections, analytical-formulation decisions, numerical transformations, statistical-model decisions, validation outcomes, accepted and rejected updates, execution checkpoints, and links between statistical input rows and their supporting evidence. An analysis object is classified as trace-complete when every analytical decision and workflow event generated during its construction is accompanied by the corresponding provenance and control record. 
Decision-trace completeness is defined as
\begin{equation}
C_{\mathrm{decision\text{-}trace}}
=
\frac{
\sum_{u=1}^{N_U}
\mathbb{I}
\left[
u \text{ is decision-trace complete}
\right]
}{
N_U
},
\label{eq:decision_trace_completeness}
\end{equation}
where \(N_U\) is the number of eligible analysis objects. An analysis object is decision-trace complete only when all decisions and workflow events generated during its construction can be traced to the relevant EAKR component, supporting evidence, and recorded validation or control state.

\textbf{Run-to-run stability} evaluates whether repeated executions produce the same synthesis decisions and aggregate evaluation results. For synthesis unit $i$ and categorical decision metric $m$, pairwise stability across $R$ runs is defined as
\begin{equation}
S_i^{(m)}
=
\frac{2}{R(R-1)}
\sum_{1\le r<s\le R}
\mathbb{I}
\left[
y_{ir}^{(m)}=y_{is}^{(m)}
\right],
\label{eq:pairwise_stability}
\end{equation}

where $y_{ir}^{(m)}$ is the value of categorical decision metric $m$ for synthesis unit \(i\) in run $r$. Overall stability for metric \(m\) is defined as
\begin{equation}
\overline{S}^{(m)}
=
\frac{1}{N_m}
\sum_{i=1}^{N_m}
S_i^{(m)},
\label{eq:overall_pairwise_stability}
\end{equation}
where \(N_m\) is the number of synthesis units eligible for metric \(m\). With five repeated executions, each eligible synthesis unit contributes ten pairwise run comparisons.

For aggregate numerical or percentage-based evaluation metrics, we also report the maximum absolute deviation from the predefined primary execution:
\begin{equation}
\Delta_m^{\max}
=
\max_{r\neq r_{\mathrm{primary}}}
\left|
A_m^{(r)}-A_m^{(\mathrm{primary})}
\right|,
\label{eq:run_deviation}
\end{equation}

where $A_m^{(r)}$ is the aggregate value of metric $m$ in run $r$. For percentage-based metrics, $\Delta_m^{\max}$ is reported in percentage points; for numerical metrics, it is reported on the corresponding metric scale.

\paragraph{\textbf{EQ4: System-level ablation.}}
\textit{How does the integrated EAKR-centred MetaSynDec system compare with direct synthesis-plan generation?}

The system-level ablation evaluates whether the complete EAKR-centred construction workflow improves the fidelity of the generated synthesis specification relative to direct generation. Both conditions received the same structured study records, review-question specification, underlying language model, and decoding parameters but differ in how the synthesis specification is constructed. MetaSynDec constructs and validates explicit analysis objects before deterministic execution, whereas the baseline directly generates a synthesis plan and statistical result without constructing the intermediate EAKR state.

Because the execution-readiness contract in Eq.~\eqref{eq:ready_predicate} is defined over explicit MetaSynDec analysis objects, it cannot be applied symmetrically to the direct-generation baseline. The system-level comparison therefore follows a hierarchical comparability protocol. It first establishes structural comparability by assessing whether each generated output addresses the same outcome, intervention--comparator contrast, and measurement-time organisation as the reference synthesis. For structurally comparable outputs, it then assesses agreement in analytical formulation, including the numerical data form, effect measure, and statistical-model policy. Categorical-result agreement and numerical-result fidelity are evaluated only when the preceding structural and formulation requirements are satisfied.

The ablation therefore evaluates the complete architectural hypothesis of MetaSynDec: analytical knowledge is constructed as an explicit, inspectable state before deterministic statistical execution. The two conditions differ in representation construction, state transitions, tool-mediated transformations, validation contracts, and iterative correction. Accordingly, the experiment estimates the value of the integrated design as a whole rather than the isolated contribution of the EAKR representation. The direct-generation prompt is provided in Appendix.

\paragraph{\textbf{Aggregation and sensitivity analyses.}}
Evaluation metrics under EQ1--EQ4 are calculated at the synthesis-unit or analysis-object level, as appropriate, and summarised across review contexts using the aggregation procedures described below. Binary agreement metrics and the evidence-set Jaccard coefficient are represented by a unit-level score \(x_{ijm} \in [0,1]\) for metric \(m\), unit or analysis object \(i\), and review context \(j\). For binary metrics, \(x_{ijm} \in \{0,1\}\), with 1 indicating that the prespecified criterion was satisfied. For evidence-set overlap, \(x_{ijm}\) is the Jaccard coefficient for the corresponding analysis object. Because the number of eligible units varies across review contexts, these metrics are summarised using both micro- and macro-averages.

For metric \(m\), the micro-average is
\begin{equation}
A_m^{\mathrm{micro}}
=
\frac{
\sum_{j=1}^{J}
\sum_{i=1}^{n_{jm}}
x_{ijm}
}{
\sum_{j=1}^{J}
n_{jm}
},
\label{eq:micro_average}
\end{equation}
where \(n_{jm}\) is the number of eligible units or analysis objects for metric \(m\) in review context \(j\), and \(J=6\) is the number of evaluation units. For binary metrics, the micro-average is the proportion of all eligible units satisfying the criterion. For the Jaccard coefficient, it is the arithmetic mean across all eligible analysis objects.

The macro-average gives equal weight to each review context containing at least one eligible unit:
\begin{equation}
A_m^{\mathrm{macro}}
=
\frac{1}{J_m}
\sum_{\substack{j=1 \\ n_{jm}>0}}^{J}
\left(
\frac{1}{n_{jm}}
\sum_{i=1}^{n_{jm}}
x_{ijm}
\right),
\label{eq:macro_average}
\end{equation}
where \(J_m\) is the number of review contexts containing at least one eligible unit or analysis object for metric \(m\). The macro-average first calculates the mean metric value within each review context and then averages these case-level values, thereby assigning equal weight to each eligible review context.

Sensitivity to individual review contexts is evaluated using leave-one-case-out (LOCO) analysis. For omitted case $j$, the micro-average is recalculated over the remaining units:
\begin{equation}
A_m^{(-j)}
=
\frac{
\sum_{\substack{k=1\\k\neq j}}^{J}
\sum_{i=1}^{n_{km}}a_{ikm}
}{
\sum_{\substack{k=1\\k\neq j}}^{J}n_{km}
}.
\label{eq:leave_one_case_out}
\end{equation}
Sensitivity is summarised by
$\left[\min_{j=1,\ldots,J} A_{m}^{(-j)},\;
\max_{j=1,\ldots,J} A_{m}^{(-j)}\right]$ 
across the six omissions.

\subsection{Implementation Details}
The agent-assisted components were implemented using Mistral-Small-3.2-24B-Instruct-2506 \cite{mistral2025small32} through the OpenRouter API. The model was selected to evaluate the proposed EAKR-centred architecture in a low-cost, accessible, and tool-capable deployment setting rather than to maximise performance through frontier-model capacity. Its support for structured output and function calling enabled its integration into the agentic workflow.
The design of MetaSynDec was grounded in established meta-analytic principles \cite{cooper_research_2017,egger_systematic_2008,hedges_statistical_2014} rather than derived from the development data. These principles informed the EAKR schema, outcome-typing rules, analytical planning logic, readiness predicates, validation contracts, and statistical decision policies. The LLM-extracted structured study records from MA1 from our study \cite{li_dataextraction} were used solely as the implementation development case for constructing, integrating, and debugging the end-to-end workflow. They were selected because they reflected the complexity of a realistic automated pipeline, including incomplete fields, inconsistent terminology, heterogeneous representations, and residual extraction uncertainty. No manually extracted study records were used to design, tune, or refine the analytical logic of the system.
After implementation had been completed, one synthesis unit from MA5, based on manually extracted study records, was used solely as an implementation-level smoke test to confirm that the frozen workflow could execute on a different outcome configuration and input source. The smoke test did not result in any changes to the EAKR schema, prompts, outcome-typing rules, planning logic, readiness predicates, validation contracts, statistical decision policies, or software implementation.

Before the reported evaluation, the complete workflow configuration was frozen, including the decoding parameters (temperature = 0.0 and top\_p = 1.0). For each synthesis unit, the first post-freeze execution was used for the primary performance analysis, followed by four additional executions for run-to-run stability assessment. Generated artifacts and intermediate states were retained only within the execution in which they were created to support checkpointing and resumption, and no generated information was transferred across runs or meta-analysis cases. All experiments were conducted between February and May 2026 without external web search, online evidence retrieval, or retrieval-augmented generation.

\section{Results}
This section reports the empirical evaluation of MetaSynDec. We first present the main evaluation results for construction effectiveness, representation and publication fidelity, and traceability, autonomy, and robustness (EQ1--EQ3). We then report the system-level ablation against direct synthesis-plan generation (EQ4), followed by an analysis of representative discrepancies between MetaSynDec and the publication references.

\subsection{Main Evaluation Results}
Table \ref{tab:main_results} summarises the primary results for EQ1--EQ3. The following subsections present the results for construction effectiveness, representation and publication fidelity, and traceability, autonomy, and robustness, respectively.

{
\begin{table}[htbp]
\centering
\caption{Summary of the main evaluation results for EQ1--EQ3.}
\label{tab:main_results}
\fontsize{7pt}{9pt}\selectfont
\setlength{\tabcolsep}{5pt}
\begin{tabular}{c l c c}
\toprule
\textbf{EQ}
& \textbf{Summary metric}
& \textbf{Eligible objects}
& \textbf{Primary result} \\
\midrule

\multirow{4}{*}{EQ1}
& Schema- and contract-valid EAKRs
& 58 synthesis units
& 58/58 (100.0\%) \\

& Full workflow completion
& 58 synthesis units
& 57/58 (98.3\%) \\

& Final supported non-pooling disposition
& 58 synthesis units
& 1/58 (1.7\%) \\

& Completion within the predefined automated-attempt bound
& 58 synthesis units
& 52/58 (89.7\%) \\

& Reviewer-authorised action
& 58 synthesis units
& 5/58 (8.6\%) \\

\midrule

\multirow{4}{*}{EQ2}
& Complete outcome-knowledge fidelity
& 37 outcomes
& 37/37 (100.0\%) \\

& Complete analysis-object fidelity
& 56 synthesis units
& 38/56 (67.9\%) \\

& Numerical result fidelity
& 55 synthesis units
& \makecell{CI overlap: 54/55 (98.2\%)\\Median D=0.26 (IQR 0.09--0.60)}\\

\midrule

\multirow{3}{*}{EQ3}
& Decision-level traceability
& 58 synthesis units
& 58/58 (100.0\%) \\

& Five-run stability
& 58 synthesis units
& \makecell{100\% stability for execution, effect measure and model policy;
\\99.3\% for significance decisions;
\\97.2\% for joint interpretation} \\

\bottomrule
\end{tabular}

\vspace{2mm}
\begin{minipage}{0.97\linewidth}
\footnotesize
\textit{Note.} EQ1 (Construction effectiveness) is evaluated at the synthesis-unit level. EQ2 (Analytical-knowledge and result fidelity) is evaluated at the analysis-object level. EQ3 (Traceability and robustness) is evaluated over workflow execution records.
\end{minipage}
\end{table}
}

\paragraph{\textbf{EQ1: Construction effectiveness.}}
MetaSynDec constructed schema- and contract-valid EAKRs for all 58 predefined synthesis units in the primary evaluation. Fifty-seven units (98.3\%) contained analysis objects eligible for quantitative pooling. One unit (1.7\%) reached a final non-pooled outcome because only one compatible study remained after evidence selection. For this unit, the eligible evidence, study exclusions, non-pooling rationale, and supporting provenance were represented explicitly in the EAKR. The 57 pooling-eligible units proceeded through deterministic statistical execution and completed the subsequent workflow stages. The full workflow-completion rate was therefore 57/58 (98.3\%). The non-pooled unit did not generate a statistical result and consequently did not produce the downstream result-dependent outputs in the current implementation. 
Of the 58 synthesis units, 52 (89.7\%) reached a final analytical state satisfying the applicable schema and contract constraints within the allowed number of automated attempts, whereas 5 (8.6\%) required reviewer-authorised action. No unit terminated with an unresolved blocking condition.

\paragraph{\textbf{EQ2: Analytical-knowledge and result fidelity.}}

Table~\ref{tab:eq2_fidelity} reports fidelity at the outcome-knowledge, analysis-object, and statistical-result levels. For the outcome knowledge represented in \(O\), MetaSynDec reproduced the reference outcome classification, outcome type, and measurement-scale and unit-harmonisation decisions for all 37 distinct review-level outcomes (37/37, 100.0\% for each metric). Complete outcome-knowledge fidelity was therefore achieved for all 37 outcomes. These results indicate that the outcome-level semantic and standardisation knowledge required for subsequent synthesis planning was preserved across the evaluated review contexts.

At the analysis-object level, MetaSynDec exactly identified the reference evidence set for 42 of the 56 objects for which the evidence set could be reconstructed (75.0\%). Figure~\ref{fig:fidelity_distributions}A further shows the unit-level distribution of evidence-set similarity across the predefined synthesis-complexity strata. Agreement was complete for both measurement-time selection and analytical formulation across all 58 analysis objects (100\%). MetaSynDec also showed high agreement for contrast assignment, effect-measure selection, and statistical-model- policy, with statistical-model-policy agreement reaching 89.5\% (51/57). Overall, MetaSynDec achieved consistently high agreement across analytical specifications, whereas agreement for exact evidence-set reconstruction was comparatively lower.

At the statistical-result level, categorical result fidelity remained high for the individual decision components but decreased under the stricter joint interpretation criterion. Direction-of-effect agreement was achieved for 48 of 57 synthesis units (84.2\%), and significance-decision agreement for 49 of 57 units (86.0\%). When both components were considered jointly, interpretation agreement was observed for 41 of 57 synthesis units (71.9\%).
Among the 55 synthesis units with directly comparable pooled estimates, numerical fidelity was high. The generated and published confidence intervals overlapped in 54 units (98.2\%), and the median confidence-interval-scaled deviation was 0.26 (IQR 0.09--0.60) and 48 of 55 units (87.3\%) differed from the published estimate by no more than one published confidence-interval half-width (Figure~\ref{fig:fidelity_distributions}B). Overall, the generated pooled estimates were numerically close to the published results relative to the uncertainty of the reference estimates.

{
\begin{table}[htbp]
\centering
\caption{Analytical-knowledge and result fidelity of the constructed EAKRs.}
\label{tab:eq2_fidelity}
\fontsize{7pt}{9pt}\selectfont
\setlength{\tabcolsep}{4pt}
\begin{tabular}{lp{4.5cm}cccc}
\toprule
\textbf{Evaluation level}
& \textbf{Metric}
& \textbf{Eligible units}
& \textbf{Overall result}
& \textbf{Macro average} 
& \textbf{LOCO sensitivity range}\\
\midrule

\multirow{4}{*}{Outcome knowledge \(O\)}
& Outcome-classification agreement
& 37
& 37/37 (100.0\%)
& 100.0\%
& 100.0--100.0\% \\
& Outcome-type agreement
& 37
& 37/37 (100.0\%)
& 100.0\%
& 100.0--100.0\% \\
& Measurement-scale and unit-harmonisation agreement
& 37
& 37/37 (100.0\%)
& 100.0\%
& 100.0--100.0\% \\
& Complete outcome-knowledge fidelity
& 37
& 37/37 (100.0\%)
& 100.0\%
& 100.0--100.0\% \\

\midrule

\multirow{8}{*}{Analysis objects \(U\)}
& Exact evidence-selection agreement (\(I_u\))
& 56
& 42/56 (75.0\%)
& 90.1\%
& 71.4--90.5\% \\

& Evidence-set Jaccard similarity
& 56
& 0.909
& 0.964 
& 0.898--0.967 \\

& Contrast-assignment agreement (\(P_u\))
& 58
& 53/58 (91.4\%)
& 96.2\%
& 90.2–95.7\%\\

& Measurement-time agreement (\(T_u\))
& 58
& 58/58 (100.0\%)
& 100.0\%
& 100.0--100.0\% \\

& Analysis-formulation agreement (\(F_u\))
& 58
& 58/58 (100.0\%)
& 100.0\%
& 100.0--100.0\% \\

& Effect-measure agreement (\(M_u^{e}\))
& 57
& 57/57 (100\%)
& 100.0\%
& 100.0--100.0\% \\

& Statistical model-policy agreement (\(M_u^{m}\))
& 57
& 52/57 (91.2\%)
& 90.7\%
& 73.9–97.9\% \\

& Complete analysis-object fidelity
& 56
& 38/56 (67.9\%)
& 81.8\%
& 63.3\% -- 75.0\% \\

\midrule

\multirow{3}{*}{Categorical result fidelity}
& Direction-of-effect agreement
& 57
& 48/57 (84.2\%)
& 93.7\%
& 81.3–95.7\% \\
& Significance-decision agreement
& 57
& 49/57 (86.0\%)
& 87.7\%
& 82.6–88.5\% \\
& Joint interpretation agreement
& 57
& 41/57 (71.9\%)
& 81.9\%
& 70.0–78.3\% \\

\midrule

\multirow{2}{*}{Numerical result fidelity}
& Confidence-interval overlap
& 55
& 54/55 (98.2\%)
& --
& -- \\
& Median CI-scaled deviation
& 55
& 0.26 (IQR 0.09--0.60)
& --
& -- \\

\bottomrule
\end{tabular}

\vspace{2mm}
\begin{minipage}{0.97\linewidth}
\footnotesize
\textit{Note.} Percentages are calculated using the eligible denominator shown for each metric. 
\end{minipage}
\end{table}
}

{
\begin{figure}[htbp]
\centering
\includegraphics[width=\linewidth]{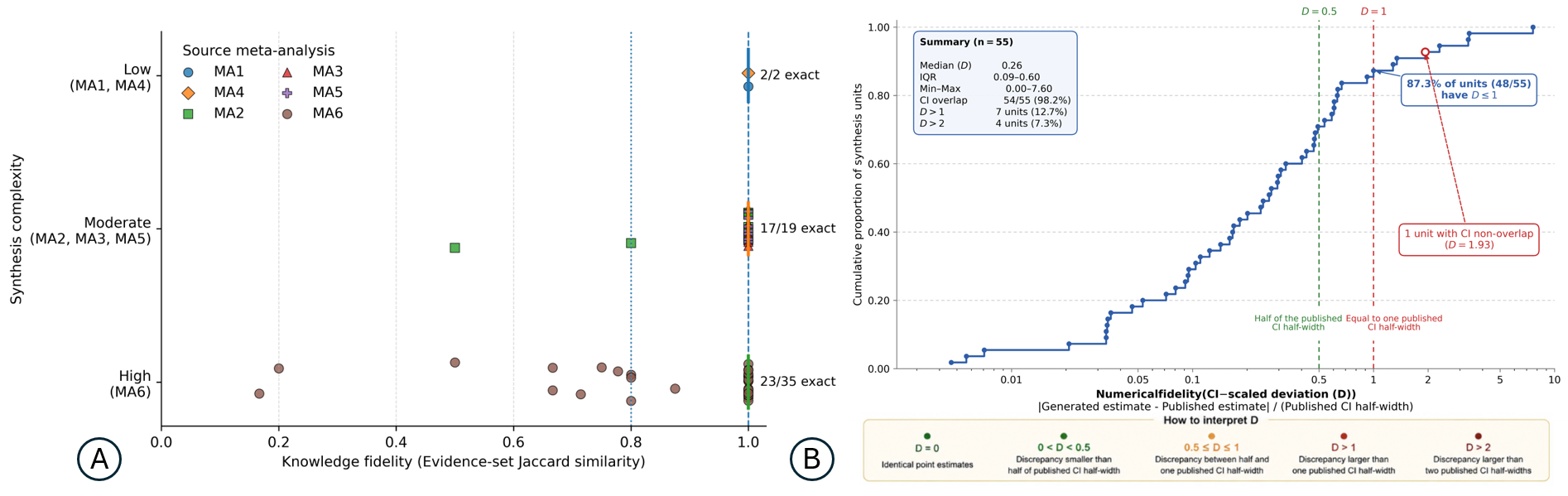}
\caption{Distribution of knowledge and numerical fidelity across synthesis units.
(A) Evidence-set Jaccard similarity for individual analysis objects, stratified by the synthesis-complexity category assigned to the source meta-analysis. Each point represents one analysis object, and marker shapes identify the source meta-analysis. Vertical reference lines indicate a Jaccard similarity of \(0.8\) and exact evidence-set agreement (\(J=1\)); labels report the number of analysis objects achieving exact agreement within each stratum.
(B) Empirical cumulative distribution of the CI-scaled deviation \(D\), defined as the absolute difference between the generated and published pooled estimates divided by the half-width of the published confidence interval. Lower values indicate closer numerical agreement. Reference lines at \(D=0.5\) and \(D=1\) indicate deviations equal to one-half and one published confidence-interval half-width, respectively.}
\label{fig:fidelity_distributions}
\end{figure}
}

\paragraph{\textbf{EQ3: Traceability and robustness.}}
All 58 eligible analysis objects were decision-trace complete, yielding a decision-trace completeness of 100.0\% (58/58). For each analysis object, the decisions generated during EAKR construction were explicitly recorded together with their supporting evidence, construction context, validation state, and subsequent workflow record. This enabled each analysis object, and each statistical result where one was produced, to be traced through the analytical chain from evidence selection and synthesis planning to the final execution-ready representation.

Run-to-run robustness was high across the five repeated executions. Successful synthesis execution, effect-measure agreement, statistical-model-policy agreement, and confidence-interval overlap were fully stable, with identical aggregate values across all five runs. Significance-decision agreement showed a pairwise stability of 99.3\%, with a \(\Delta\) of 1.8 percentage points from the primary-run value. Joint interpretation agreement showed slightly greater variation, with a pairwise stability of 97.2\% and \(\Delta\) of 3.5 percentage points. The median confidence-interval-scaled deviation ranged from 0.26 to 0.32 across the five runs.
These results show that the constructed EAKRs preserved complete decision-level traceability. The repeated executions also produced highly stable, although not fully deterministic, analytical and statistical outputs.

\subsection{System-Level Ablation against Direct Generation}
\label{sec:system_ablation}
Table~\ref{tab:ablation} reports the system-level ablation results according to the hierarchical comparability protocol defined under EQ4. Results are presented sequentially for reference synthesis-structure agreement, analysis-formulation agreement, categorical-result agreement, and numerical-result fidelity, with each downstream level evaluated only when the preceding comparability requirements were satisfied. For both conditions, the prospectively designated first execution after configuration freeze was used for the primary analysis, regardless of completion status. The four additional direct-generation executions were analysed separately to assess run-to-run stability.

\paragraph{\textbf{Primary results.}}
Reference synthesis-structure agreement was observed for 57 of the 58 synthesis units under MetaSynDec and for 23 units under direct generation (98.3\% versus 39.7\%, exact McNemar $p<0.001$). The lower agreement under direct generation primarily reflected departures from the specified measurement-time organisation. Despite an explicit prompt requirement for time-specific synthesis, direct generation produced 37 outputs for the 58 reference synthesis units and frequently combined multiple measurement-time-specific targets into broader analyses. These outputs therefore could not be mapped one-to-one to the predefined synthesis units and were classified as structurally non-comparable.

Analysis-formulation agreement was subsequently evaluated for the 23 synthesis units satisfying the structural-comparability criterion under both conditions. MetaSynDec agreed with the reference analytical formulation for all 23 units, whereas direct generation agreed for one unit (100.0\% versus 4.3\%, exact McNemar $p<0.001$). Direct generation used endpoint values for all 23 structurally comparable units, despite the prompt explicitly requiring the numerical data form to be selected according to the available study data and the specified synthesis target. 

Categorical-result agreement and numerical-result fidelity were evaluated only for synthesis units satisfying both structural and analytical comparability. Because only one direct-generation output met these prerequisites, aggregate comparison of these downstream metrics was not informative. The observed limitation therefore arose primarily from divergence in synthesis specification.

\paragraph{\textbf{Sensitivity results.}}
After exclusion of MA6, all 23 remaining reference synthesis units were structurally comparable under both conditions. MetaSynDec agreed with the reference analytical formulation for all 23 units, whereas direct generation agreed for only one unit (100.0\% versus 4.3\%). This analysis indicates that the structural difference in the full comparison was concentrated in the multiple-measurement-time case, while the formulation difference remained evident among the single-time synthesis units.

\paragraph{\textbf{Run-to-run stability.}}
Across the five direct-generation executions, reference synthesis-structure agreement remained unchanged at 23 of 58 units (39.7\%) in every run. Among these 23 structurally comparable units, only one satisfied the analysis-formulation criterion in each execution. The repeated executions therefore showed no run-to-run variation in either structural agreement or formulation agreement.

{
\begin{table}[htbp]
\centering
\caption{Workflow ablation results (EQ4) comparing MetaSynDec with the direct-generation baseline.}
\label{tab:ablation}
\fontsize{8pt}{10pt}\selectfont
\setlength{\tabcolsep}{6pt}
\begin{tabular}{lcccc}
\toprule
\textbf{Metric} 
& \textbf{Baseline}
& \textbf{MetaSynDec}
& \textbf{$\Delta$ (pp)}
& \textbf{$p$} \\
\midrule

Synthesis-structure agreement
& 23/58 (39.7\%)
& 57/58 (98.3\%)
& \textbf{+58.6}
& $<0.001$ \\

Analysis-formulation agreement$^\dagger$
& 1/23 (4.3\%)
& 23/23 (100.0\%)
& \textbf{+95.7}
& $<0.001$ \\

\bottomrule
\end{tabular}

\vspace{1.5mm}
\begin{minipage}{0.92\linewidth}
\footnotesize
\textit{Note.} $^\dagger$ Calculated on synthesis units matched by the direct-generation baseline, enabling paired comparison between methods.
The direct-generation baseline was stable across five independent runs: successful synthesis execution was 23/58 in all runs, and formulation agreement was 1/23 in all runs.
$\Delta$ indicates the difference between MetaSynDec and the baseline in percentage points.
\end{minipage}
\end{table}
}

\subsection{Analysis of Discrepancies}
To characterise the limitations of MetaSynDec as a knowledge-based synthesis system, we conducted a manual analysis of the 17 synthesis units for which the generated and published results differed in their joint direction-and-significance classification. Each case was examined as a synthesis-construction discrepancy rather than solely as an output mismatch. We inspected the structured evidence records available to MetaSynDec, the corresponding source-study reports, the constructed EAKR, and the published meta-analysis. Each disagreement was assigned to the earliest stage at which the knowledge required to construct or execute the reference synthesis became unavailable, insufficiently grounded, or inconsistent with the published result. 
The classification was conducted by L.L., and the assigned categories were reviewed through discussion until consensus was reached. The categories constitute a qualitative error taxonomy intended to explain system behaviour rather than mutually independent statistical error classes. Figure~\ref{fig:disagreement_sources} summarises the resulting sources of disagreement.

\paragraph{\textbf{Knowledge-coverage gaps.}} The largest category of disagreement were knowledge-coverage gaps, accounting for 8 of the 17 disagreements. In these units, synthesis-relevant measurements were reported in the source-study materials but were absent from the structured evidence records supplied to MetaSynDec. The inherited evidence dataset primarily represented outcomes reported in the main article text and did not consistently include measurements available only in supplementary materials. Consequently, the evidence set available for constructing \(I_u\) and populating the corresponding numerical inputs in \(D_u\) was incomplete relative to that used in the published synthesis.
These disagreements therefore originated upstream of analysis-object construction and do not represent failures of downstream analytical reasoning. Most occurred in MA6, where relevant outcome information was distributed between the main reports and supplementary materials. For several synthesis units, MetaSynDec and the published meta-analysis consequently operated over different evidence states. This finding indicates that the fidelity of automated synthesis depends not only on the analytical reasoning process, but also on the coverage of the structured knowledge base from which the EAKR is constructed.

\paragraph{\textbf{Semantic-inference failures.}} The second largest category comprised semantic-inference failures, accounting for 5 of the 17 disagreements. In these units, the required numerical evidence was present in the structured evidence records, but MetaSynDec could not reliably bind those records to the analytical roles required by the reference synthesis. Typical examples included unresolved intervention--comparator assignments, ambiguous study-arm roles, and temporal descriptions such as ``latest follow-up'' that could not be mapped unambiguously to a target measurement time.
Within the EAKR, these units primarily affected construction of the contrast \(P_u\), the measurement-time selection \(T_u\), or the association between the selected evidence records \(I_u\) and the target analysis object. Where the ambiguity could not be resolved from the available evidence, it was retained as an unresolved issue in \(Q_u\) rather than converted into an unsupported analytical decision.
A representative case occurred in an RCT from MA2. The structured evidence contained the required outcome values and summary statistics, but the available contextual information was insufficient for the agent to resolve the intervention--comparator orientation. As a result, the corresponding evidence record could not be assigned reliably to the contrast \(P_u\), and the study was not included in the executable synthesis. The published review, by contrast, assigned the outcome to the intervention arm and included it in the analysis. The disagreement therefore originated from semantic inference rather than from missing numerical evidence.

\paragraph{\textbf{Reference-synthesis reproducibility discrepancies.}} The remaining four disagreements were classified as reference-synthesis reproducibility discrepancies. In these units, the structured numerical inputs in \(D_u\) were consistent with the statistical quantities explicitly reported in the source studies, but executing the resulting analysis object did not reproduce the published synthesis result. Manual inspection indicated that several source studies labelled their dispersion measures as standard errors, whereas reproducing the published meta-analysis required treating those values as standard deviations.
A representative case occurred in MA3. The source reports explicitly identified the relevant dispersion quantities as standard errors. Preserving those stated statistical types in \(D_u\) produced a pooled estimate and confidence interval that differed from the published result. The reference result could be reproduced only by reinterpreting the reported standard errors as standard deviations. The same pattern was observed in three additional synthesis units.
These units do not indicate that the constructed EAKR lacked the information needed for execution. Rather, they expose a conflict between the statistical semantics stated in the source evidence and the unstated assumptions required to reproduce the published synthesis. From a knowledge-based systems perspective, they illustrate that disagreement with a published reference may arise even when the system preserves the source-reported statistical types correctly.

{
\begin{figure}[htbp]
\centering
\includegraphics[width=\linewidth]{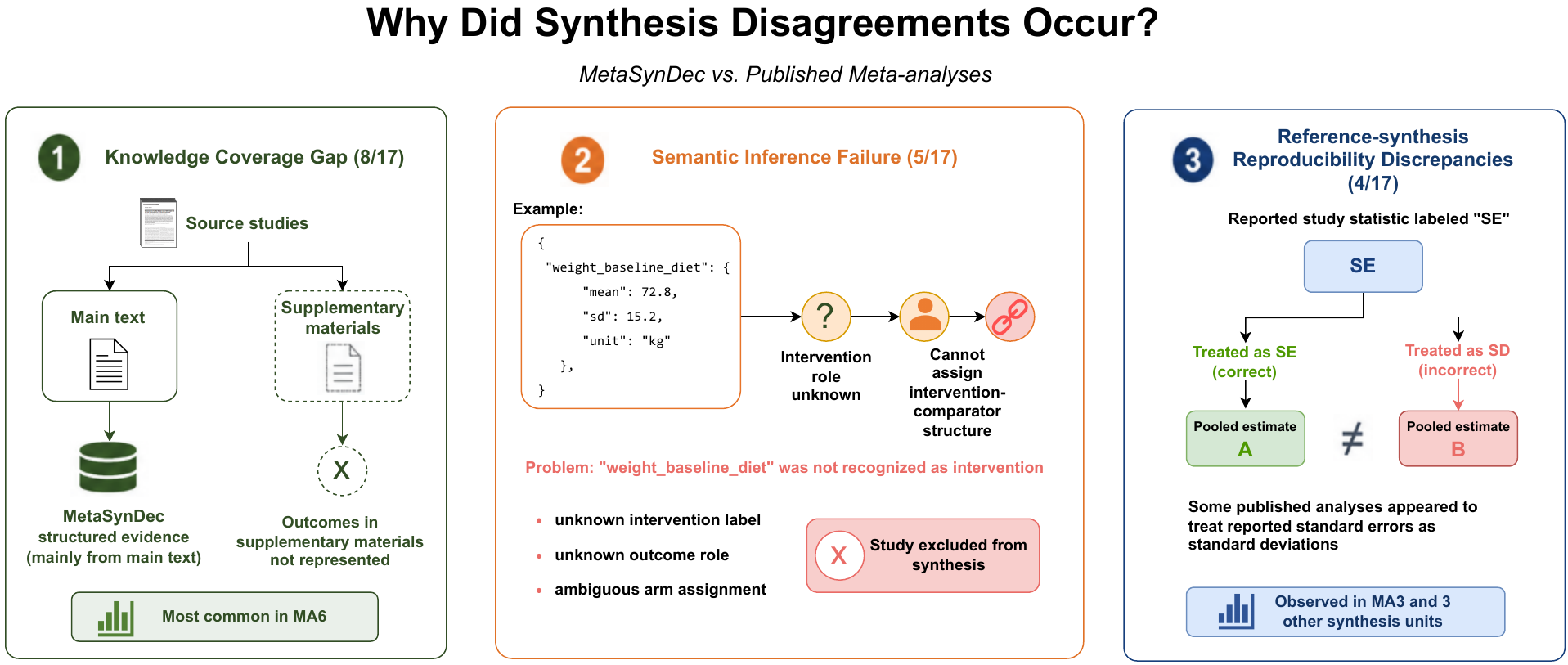}
\caption{Diagnosed sources of synthesis disagreement. Among the 17
disagreement units examined, eight reflected knowledge-coverage gaps,
five reflected semantic-grounding failures, and four reflected
reference-synthesis reproducibility discrepancies. The examples
illustrate how explicit EAKR components enabled each disagreement to be
localised to evidence coverage, analytical interpretation, or the
published reference synthesis.}
\label{fig:disagreement_sources}
\end{figure}
}

\section{Discussion}
This study investigated whether meta-analysis synthesis can be formulated as a knowledge-based computational process that transforms structured study evidence into executable analytical specifications. The evaluation addressed four complementary questions: whether executable analytical knowledge can be constructed (EQ1), whether the resulting knowledge is analytically faithful to published syntheses (EQ2), whether it remains traceable and reproducible throughout execution (EQ3), and and how the integrated EAKR-centred system compares with direct synthesis-plan generation (EQ4). Together, these findings suggest that reliable automated meta-analysis depends not only on language-model reasoning, but also on explicit knowledge representation, constrained knowledge construction, and decision-level validation.

\subsection{EAKR as a Computational Knowledge Representation}
EQ1 examined whether structured study evidence can be transformed into an executable representation of analytical intent before statistical pooling. The results demonstrate that this transformation is feasible through the proposed EAKR, which explicitly represents the analytical decisions required to define a synthesis rather than leaving them implicit within language model reasoning.
This finding extends the notion of an explicit synthesis specification from a methodological concept to a computational knowledge representation. Unlike structured evidence records, which describe what individual studies reported, the EAKR captures the analytical decisions that determine how evidence should be organised into executable synthesis objects, including evidence membership, intervention and comparator roles, measurement-time constraints, analytical formulations, statistical policies, provenance, and unresolved issues.

From a knowledge-based perspective, the EAKR functions as a domain-specific representation that separates analytical reasoning from statistical execution. Statistical computation therefore becomes the execution of an explicitly represented knowledge state rather than the interpretation of heterogeneous study evidence. This separation enables analytical intent to become inspectable, reusable, and independently verifiable before any statistical model is executed.

\subsection{Workflow-Constrained Construction of Executable Representations}
Once analytical intent has been represented explicitly, the central systems challenge is to construct an executable representation that faithfully preserves the analytical decisions of the target review. The results for EQ2 provide evidence that MetaSynDec addresses this challenge effectively. The high levels of analytical fidelity observed across outcome knowledge, analysis-object construction, and statistical execution indicate that the proposed workflow can reliably reconstruct executable analytical knowledge from structured evidence.

These findings also clarify why workflow decomposition improves synthesis performance. The benefit does not arise merely from distributing the task across multiple language-model interactions. Rather, each workflow stage performs a constrained update to a shared analytical knowledge state. Modifications are restricted to predefined fields, stage-specific methodological requirements must be satisfied, and intermediate outputs are checked against the applicable schema and contract constraints before downstream execution can proceed. The workflow therefore transforms language-model outputs from unconstrained synthesis proposals into bounded knowledge-state transitions.
This constraint is particularly important because meta-analysis involves interpretation-intensive decisions that cannot be fully encoded through deterministic rules. These include mapping heterogeneous outcome descriptions to review-level concepts, identifying intervention and comparator roles, resolving follow-up definitions, determining whether reported measurements satisfy the requirements of a target synthesis unit, and translating these decisions into valid analysis objects. Language models are well suited to providing semantic interpretation in such units, but their outputs must remain governed by the methodological structure of the review rather than being generated through unconstrained end-to-end reasoning.

The ablation results further demonstrate the computational role of this structure. When the explicit workflow was removed, synthesis-unit matching and analytical-formulation agreement deteriorated substantially despite the use of identical structured inputs. This suggests that the workflow contributes more than procedural organization. It provides the computational mechanism through which the executable representation is constructed incrementally while preserving consistency across outcome interpretation, evidence matching, analysis-object formulation, and statistical execution.
Viewed from this perspective, MetaSynDec is not primarily an agent-orchestration framework, but a knowledge-construction system. Language models supply semantic interpretation where deterministic rules are insufficient, whereas the workflow governs how those interpretations are incorporated into a coherent, validated, and executable analytical representation. The agent is therefore responsible for constructing the representation rather than replacing it, and deterministic statistical procedures operate only after the relevant analytical knowledge has been made explicit.

\subsection{Validation and Diagnosis through Explicit Knowledge Representation}

A major consequence of representing synthesis as explicit analytical knowledge is that it changes the object of verification in automated meta-analysis. Conventional end-to-end systems are typically evaluated by comparing final statistical outputs with published meta-analysis results. Such comparisons indicate whether the final estimate is similar to the reference, but they provide little information about where disagreements originate or whether the underlying analytical decisions were constructed correctly.
The explicit representation introduced in this work enables validation to operate at the level of analytical knowledge rather than only at the level of statistical outputs. Because each analysis object records the selected evidence, intervention and comparator assignments, measurement time, analytical formulation, statistical policy, provenance, and any remaining unresolved issues, each component of the synthesis can be inspected independently before statistical execution. Consequently, verification becomes a process of examining whether the constructed knowledge faithfully represents the intended analytical specification, rather than simply asking whether the pooled estimate matches a published result.

The disagreement analysis further illustrates the diagnostic value of this representation. Rather than treating disagreements as undifferentiated model failures, the explicit analytical representation made it possible to distinguish disagreements arising from incomplete knowledge coverage, semantic-inference failures, and reference-synthesis reproducibility discrepancies. These categories reflect different knowledge conditions rather than different numerical errors. 
From a knowledge-construction perspective, these categories imply different corrective actions. Knowledge-coverage gaps motivate richer evidence acquisition, semantic-inference failures motivate improved reasoning or reviewer intervention, whereas reproducibility discrepancies require re-examination of the reference synthesis rather than modification of the constructed representation. The ability to localise disagreements in this manner is a direct consequence of making analytical knowledge explicit.

\subsection{Broader Implications for Scientific Workflow Design}
Although developed for meta-analysis synthesis, the principles underlying the EAKR are not specific to evidence synthesis. Many scientific workflows require heterogeneous evidence to be transformed into explicit analytical decisions before deterministic computation can proceed. In such settings, the central automation challenge lies not in executing the computation itself, but in constructing an explicit representation that specifies how computation should be performed \cite{leipzig2021metadata, colonnelli2025swirl}.
The findings suggest a representation-centred design pattern for LLM-assisted scientific workflows. Rather than relying on language models to generate complete analytical results directly, language models can instead contribute by constructing explicit intermediate knowledge under domain-specific constraints \cite{agarwal2024tic}. Formal schemas, methodological rules, state-transition constraints, and readiness validation then govern how this knowledge is progressively refined before deterministic computational services execute the final analysis. This separation allows probabilistic semantic reasoning and deterministic computation to complement rather than replace one another \cite{colonnelli2025swirl, agarwal2024tic}.
Although the specific structure of an EAKR is domain dependent, the underlying principle may extend to other knowledge-intensive scientific tasks, including guideline development, health technology assessment, regulatory evidence review, and evidence grading, where analytical reasoning precedes formal computation. More generally, this work suggests that reliable scientific automation may depend less on asking language models to produce complete analytical outputs and more on using them to construct explicit, executable knowledge that can be inspected, checked against predefined constraints, revised, and executed under methodological control \cite{leipzig2021metadata,pritchard2025}.

\subsection{Limitations and Future Research}
This study has several limitations. First, the evaluation used manually curated, structured study-level evidence to isolate the synthesis-specification process from upstream errors. The reported performance should therefore be interpreted within a controlled-input setting. In end-to-end evidence synthesis, errors in retrieval, screening, data extraction, or outcome normalisation may propagate into the EAKR and affect subsequent analytical decisions. Future evaluations should assess the robustness of EAKR construction under realistic upstream uncertainty and incomplete or inconsistently structured evidence.
Second, even under controlled inputs, exact evidence-set reconstruction was the least reliable component, despite high overall analytical-specification fidelity. The explicit structure of the EAKR made these failures traceable, but did not fully resolve the underlying problems of knowledge coverage and semantic inference. Future work should therefore improve retrieval from supplementary materials, develop controlled intervention and comparator ontologies, represent uncertainty explicitly, and incorporate targeted reviewer input for ambiguous evidence-selection decisions.
Third, the evaluation was restricted to pairwise aggregate-data meta-analysis and a relatively small set of review contexts. Although this setting captures common decisions involving outcome abstraction, follow-up alignment, comparison-role assignment, effect-size formulation, and model-policy selection, it does not cover the full range of evidence-synthesis designs. Network meta-analysis, diagnostic-test-accuracy meta-analysis, and individual-participant-data meta-analysis may require richer analytical objects, additional validity constraints, and different execution services.
Finally, this study evaluated one concrete instantiation of EAKR within pairwise aggregate-data meta-analysis. The findings demonstrate that, in this setting, explicit analytical objects can be constructed under predefined constraints, revised traceably, and executed deterministically; however, broader generalisability remains to be established. Extending EAKR to other forms of evidence synthesis will require task-specific schemas, constraints, and execution services built on a shared representational core. Future work should therefore focus on reusable EAKR modules and configurable methodological components, rather than redesigning the synthesis representation for each application.

\section{Conclusion}
This work introduced the Executable Analytical Knowledge Representation (EAKR) as a computational abstraction for meta-analysis synthesis. It formalises the analytical layer between structured study evidence and statistical execution by explicitly representing evidence membership, analytical roles, outcome and time-point grounding, statistical requirements, provenance, and unresolved issues.
The proposed agentic harness instantiates this formulation by using language-model agents to propose bounded updates to the EAKR, while schemas, methodological constraints, validation rules, readiness checks, and statistical input contracts govern whether those updates can proceed to execution. The system therefore illustrates a division of responsibility in which language models support semantic knowledge construction, whereas formal representations and deterministic services govern analytical admissibility and computation.
More broadly, the EAKR provides a representation-centred foundation for automated evidence synthesis. By making analytical decisions explicit before execution, it enables synthesis knowledge to be inspected, revised, and re-executed independently of the language model that helped construct it. This shifts the role of AI in meta-analysis from generating final results to constructing executable analytical knowledge under methodological control.

\bibliographystyle{unsrtnat}
\bibliography{references}

\end{document}